\documentclass[10pt,twocolumn,letterpaper]{article}
\usepackage[accsupp]{axessibility} 
\usepackage{cvpr}              

\newcommand{\change}[1]{{\color{Black}#1}}

\usepackage{amsmath}
\usepackage{amssymb}
\usepackage{mathrsfs}
\usepackage[table]{xcolor}

\definecolor{cvprblue}{rgb}{0.21,0.49,0.74}
\usepackage[pagebackref,breaklinks,colorlinks,allcolors=cvprblue]{hyperref}
\usepackage{graphicx}
\usepackage{amsmath}
\usepackage{amssymb}
\usepackage{multirow}
\usepackage{booktabs}
\usepackage{array}
\usepackage{amsfonts}
\usepackage{color}
\usepackage{xcolor}
\usepackage{algorithmicx,algpseudocode}
\usepackage{setspace}
\usepackage{stfloats}

\usepackage[linesnumbered,ruled,vlined]{algorithm2e}
\renewcommand{\paragraph}[1]{\vspace{-1.5mm}{\flushleft\textbf{#1}}}

\def\paperID{xxxx} 
\def\confName{CVPR}
\def\confYear{2026}

\begin{document}
\title{FreeScale: Scaling 3D Scenes via Certainty-Aware Free-View Generation}

\author{
Chenhan Jiang$^{1*\dagger}$
\quad Yu Chen$^{2*}$
\quad Qingwen Zhang$^{3}$ 
\quad Jifei Song$^{4}$ \\
\quad Songcen Xu$^{5}$
\quad Dit-Yan Yeung$^{1}$
\quad Jiankang Deng$^{6\dagger}$\\
\vspace{-15mm}
}


\twocolumn[{
\maketitle
\begin{center}
    \includegraphics[width=0.9\textwidth]{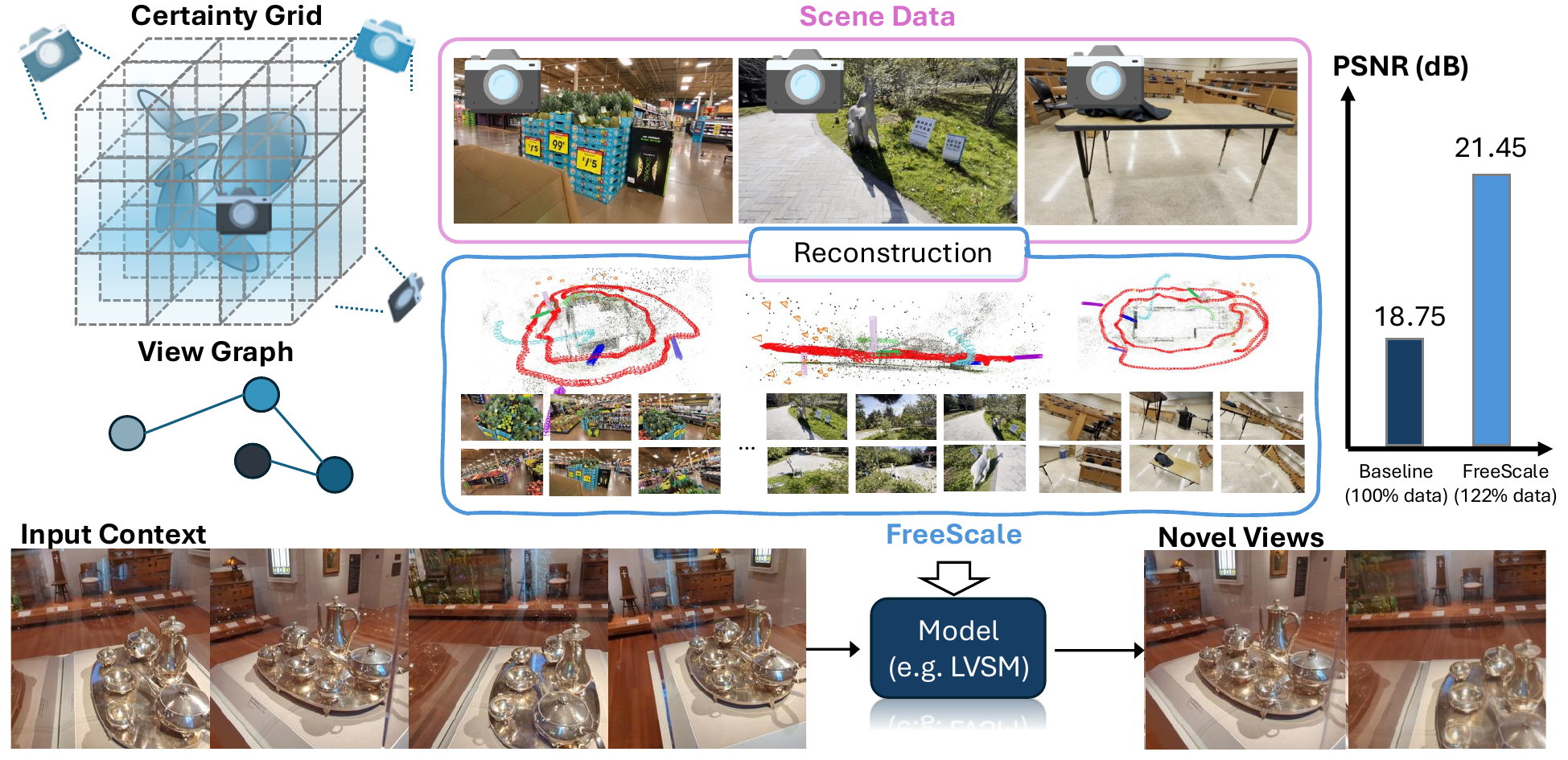}
    \vspace{-3mm}
    \captionof{figure}{%
        {\small We introduce \change{\textbf{FreeScale}}, a framework that scales current scene data by generating free-view images from reconstructed scene geometry, which can be used for feed-forward model training. Training LVSM with an additional 22\% of generated free-views significantly improves sparse-view reconstruction from PSNR 18.75 to 21.45, particularly enhancing its generalization to large camera motion.}
    }
    \label{fig: teaser}
\end{center}
}]

\newcommand\blfootnote[1]{%
\begingroup
\renewcommand\thefootnote{}\footnote{#1}%
\addtocounter{footnote}{-1}%
\endgroup
}
\blfootnote{$^*$Equal contribution. $^1$Hong Kong University of Science and Technology\hspace{2mm} $^2$National Univeristy of Singapore\hspace{2mm} $^3$KTH Royal Institute of Technology\hspace{2mm} $^4$University of Surrey\hspace{2mm} $^5$Independent Researcher\hspace{2mm} $^6$Imperial College London\hspace{2mm} 
$^\dagger$Corresponding Author: \url{jchcyan@gmail.com}, \url{j.deng16@imperial.ac.uk}
}

\begin{abstract}

The development of generalizable Novel View Synthesis (NVS) models is critically limited by the scarcity of large-scale training data featuring diverse and precise camera trajectories. While real-world captures are photorealistic, they are typically sparse and discrete. Conversely, synthetic data scales but suffers from a domain gap and often lacks realistic semantics. We introduce \change{FreeScale}, a novel framework that leverages the power of scene reconstruction to transform limited real-world image sequences into a scalable source of high-quality training data. Our key insight is that an imperfect reconstructed scene serves as a rich geometric proxy, but naively sampling from it amplifies artifacts. To this end, we propose a certainty-aware free-view sampling strategy \change{identifying} novel viewpoints that are both semantically meaningful and minimally affected by reconstruction errors. We demonstrate \change{FreeScale}'s effectiveness by scaling up the training of feedforward NVS models, achieving a \change{notable gain of 2.7 dB in PSNR} on challenging out-of-distribution benchmarks. Furthermore, we show that the generated data can actively enhance per-scene 3D Gaussian Splatting optimization, leading to consistent improvements across multiple datasets. Our work provides a practical and powerful data generation engine to overcome a fundamental bottleneck in 3D vision. \change{Project page: \url{https://mvp-ai-lab.github.io/FreeScale}}.

\end{abstract}    

\vspace{-3mm}
\section{Introduction}
\label{sec:intro}

Novel View Synthesis (NVS), which aims to generate photorealistic images from unobserved viewpoints given sparse input, is a foundational problem in computer vision and graphics. Optimization-based methods like Neural Radiance Fields (NeRF)~\cite{mildenhall2021nerf} and the highly efficient 3D Gaussian Splatting (3DGS)~\cite{kerbl20233d} have achieved impressive visual fidelity through per-scene reconstruction. More recently, the field has seen a growing trend toward generalizable feedforward models~\cite{chen2024mvsplat360,jin2024lvsm,jiang2025rayzer} that learn cross-scene priors for efficient 3D reconstruction at inference time.

Despite these advances, a critical data bottleneck persists: the scarcity of large-scale datasets with diverse and accurate camera trajectories, which limits the scalability and robustness of feedforward models, especially for large, free-viewpoint camera motions. Moreover, optimization-based methods remain sensitive to imperfect captures (e.g., insufficient scene coverage or inaccurate camera poses), which leads to noticeable geometric errors and artifacts. Furthermore, collecting large-scale, high-quality real-world data with meticulous camera calibration remains a laborious and expensive process.

Existing approaches to scale data have significant limitations. Methods leveraging simulators (e.g., Blender~\cite{hess2013blender}) can generate synthetic data with perfect camera control~\cite{raistrick2024infinigen,roberts2021hypersim,xie2024lrm,wang2020tartanair,srivastava2022behavior}, but they introduce a substantial synthetic-to-real domain gap. For instance, LRM-Zero~\cite{xie2024lrm} trained entirely on synthetic data shows significantly reduced visual fidelity compared to models trained on real data. While Megasynth~\cite{jiang2025megasynth} bypasses realistic semantics through amorphous geometry and texture stacking, it suffers from poor data efficiency. Alternatively, diffusion-based methods~\cite{nair2025scaling} can generate photorealistic images but typically fail to provide the accurate camera poses essential for NVS training.

This reveals a critical gap. Real-world captures provide the desired photorealism and semantics but exist only as discrete, sparsely sampled sequences. While one can reconstruct a continuous scene representation from them, naively sampling novel views from this reconstruction often yields images marred by artifacts or poor semantics. We propose \change{FreeScale}, as shown in Figure~\ref{fig: teaser}, a framework that \textit{not only creates a continuous representation via reconstruction but, more importantly, introduces a principled method to sample high-fidelity, semantically meaningful novel views with accurate poses} from it. Unlike methods focused solely on per-scene quality or scaling feedforward models, our approach functions as a data engine to enhance and scale existing real-world captures.
The key challenge is identifying free views that maximally capture under-constrained geometry while being minimally contaminated by reconstruction artifacts. To address this, we introduce a certainty-aware free-view sampling strategy. First, we construct a certainty grid from the reconstructed geometry to enable diverse and semantically meaningful look-at areas. Second, we build a view graph to establish geometric correspondences between generated and collected real-world data, guiding final view selection and rectification.

We conduct extensive experiments on challenging datasets containing out-of-distribution views and large camera motions. We validate \change{FreeScale} in two key applications:
\textbf{1) Scaling feedforward NVS models} by augmenting training data with our generated views, expanding the training set to 1.22$\times$ its original dataset size~\cite{ling2024dl3dv}. Our approach achieves a 2.7 dB PSNR improvement \change{over baseline methods}.
\textbf{2) Enhancing per-scene 3DGS optimization} through active non-certainty exploration, demonstrating consistent improvements on DL3DV~\cite{ling2024dl3dv}, Nerfbusters~\cite{Nerfbusters2023}, and Tanks and Temples~\cite{DBLP:journals/tog/KnapitschPZK17} datasets.
Our contributions are three-fold:
\begin{itemize}
\change{
\item A novel framework FreeScale, which leverages certainty-guided sampling to generate diverse, high-quality free-view images from sparse inputs. This expands the viewpoint coverage to  boost downstream tasks significantly.
\item A certainty-based view graph to efficiently manage and filter candidate viewpoints, ensuring the generated free-view images are both informative and photorealistic.
}
\item Comprehensive validation demonstrating significant improvements in both feedforward reconstruction and per-scene optimization.
\end{itemize}
\begin{figure*}[t]
\centering
\includegraphics[width=\linewidth]{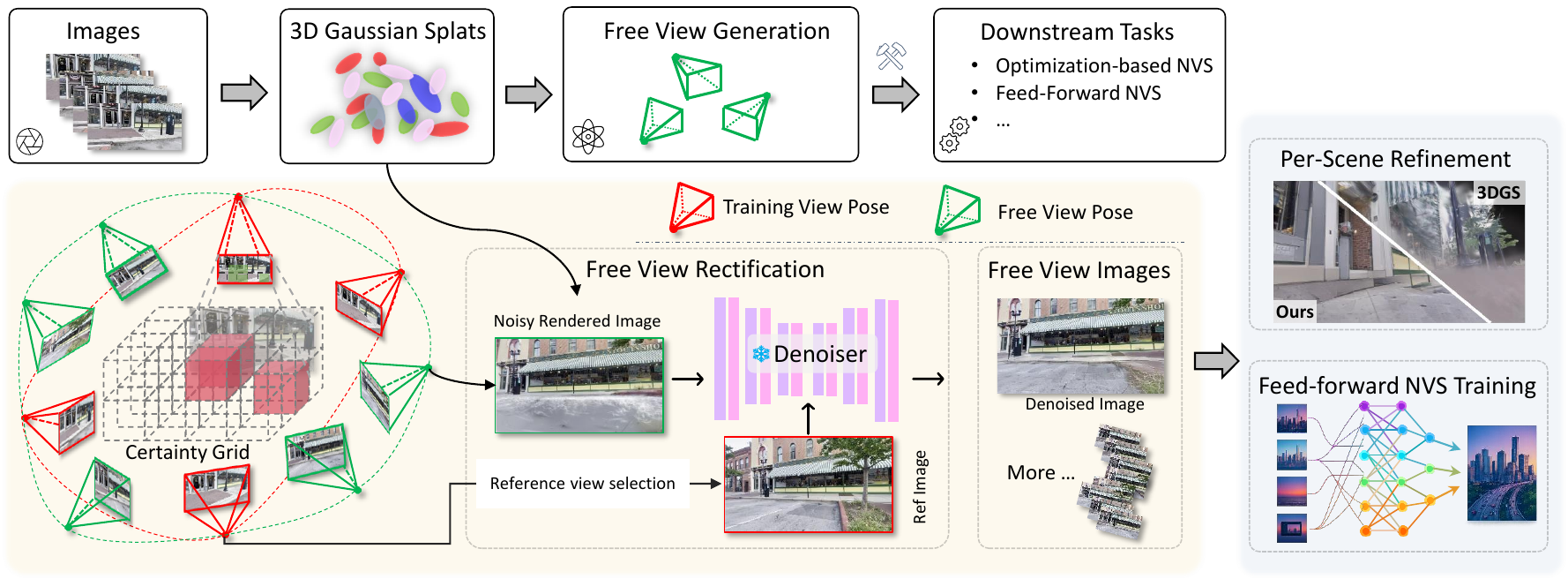}
   \vspace{-2em}
   \caption{\textbf{FreeScale generation pipeline.} Our overall pipeline consists of three phases. First, given an image sequence, we reconstruct the scene as a continuous 3D representation, which allows us to place arbitrary viewpoint candidates. Second, we perform certainty-aware free-view synthesis: we establish a view graph based on a certainty grid and filter redundant candidates. Finally, we apply image rectification to produce the final free-views. The generated data can then be used to train feed-forward models like LVSM and refine the scene Gaussians.
}
\label{fig:method}
\vspace{-0.8em}
\end{figure*}

\section{Related Works}
\label{sec:related}
\noindent\textbf{Scene-level Novel View Synthesis.}
Neural Radiance Fields (NeRF)~\cite{mildenhall2021nerf} and the highly efficient 3D Gaussian Splatting (3DGS)~\cite{kerbl20233d} are prime approaches for image synthesis at novel view points. However, these methods are susceptible to artifacts like floaters and ghosting when trained with sparse inputs or evaluated on out-of-distribution viewpoints. A significant body of work addresses this by imposing geometric constraints, using either sparse~\cite{DBLP:conf/cvpr/DengLZR22,DBLP:conf/cvpr/0007Z0ZNZ024} or dense~\cite{DBLP:conf/cvpr/RoessleBMSN22} depth supervision, or by developing sophisticated strategies to control the placement and movement of Gaussian primitives~\cite{DBLP:conf/eccv/ZhangHLHZ24,DBLP:conf/eccv/LiLDZLT24,DBLP:conf/cvpr/0005YXX0L024,DBLP:conf/eccv/VerverasPSDZ24}. Other approaches mitigate appearance variations by decoupling view-dependent effects using appearance embeddings~\cite{DBLP:conf/nips/XuMP24, DBLP:conf/cvpr/LinLTLLLLWXYY24, DBLP:conf/nips/ChenL24} or bilateral grids~\cite{DBLP:journals/corr/abs-2506-05280, DBLP:journals/tog/WangWGX24}.

Despite their high quality, these per-scene optimization methods are computationally expensive and do not generalize. This limitation has spurred the development of \emph{generalizable, feed-forward} models. These approaches learn cross-scene priors from large datasets, typically by constructing cost volumes to reason about geometry~\cite{DBLP:conf/eccv/ChenXZZPGCC24, DBLP:conf/eccv/LiuWHSYZCLL24, chen2024mvsplat360, DBLP:conf/cvpr/CharatanLTS24, DBLP:conf/cvpr/WangWGSZBMSF21, DBLP:conf/cvpr/ChenL23} or by leveraging powerful 3D foundation models (\eg, DUSt3R~\cite{DBLP:conf/cvpr/Wang0CCR24}, VGGT~\cite{DBLP:conf/cvpr/WangCKV0N25}) for improved initialization and generalization~\cite{DBLP:conf/cvpr/XuPWBB0P25, DBLP:conf/iclr/YeLXLP0P25, DBLP:journals/corr/abs-2505-23716}. Transformer-based architectures have also been adapted for this task, operating either on 3D Gaussians~\cite{DBLP:conf/eccv/ZhangBTXZSX24, DBLP:journals/corr/abs-2410-12781} or directly in 2D~\cite{jin2024lvsm,jiang2025rayzer}.

A critical bottleneck for these generalizable models is the scarcity of large-scale, diverse training data with accurate and extensive camera trajectories. This data limitation caps their scalability and robustness, particularly for large, free-viewpoint camera motions. Furthermore, the performance of \emph{both} optimization-based and feed-forward methods is highly sensitive to input imperfections, such as inaccurate camera poses.

\noindent\textbf{Scaling up Scene-level Data and Priors.}
The performance of generalizable NVS models is fundamentally constrained by the scale and quality of training data. This challenge has been addressed through several distinct paradigms.

One paradigm seeks to \emph{generate synthetic data} at scale. This includes using 3D simulators like Blender to produce data with perfect camera control but a significant synthetic-to-real domain gap~\cite{raistrick2024infinigen, roberts2021hypersim, xie2024lrm, wang2020tartanair, srivastava2022behavior}. Methods like Megasynth~\cite{jiang2025megasynth} expand diversity through amorphous geometry, but this often leads to poor data efficiency. A more recent approach leverages powerful 2D diffusion models to generate photorealistic images~\cite{nair2025scaling}. However, a fundamental limitation of these \emph{data-generation} methods is their inability to provide accurate, consistent camera poses and multi-view coherent geometry, which are essential for training robust 3D reconstruction systems.

Another paradigm uses pre-trained models as \emph{priors to enhance} per-scene optimization. Methods distill 2D diffusion priors to mitigate sparse-view artifacts~\cite{DBLP:conf/iccv/WarburgWTHK23, DBLP:journals/corr/abs-2311-15127}, such as ReconFusion~\cite{DBLP:conf/cvpr/WuMHPGWSVBPH24} generating pseudo-labels for NeRFs. For 3DGS, 3DGS-Enhancer~\cite{liu20243dgs} and DIFIX3D~\cite{wu2025difix3d+} refine pseudo-views and correct artifacts. While effective for individual scenes, these \emph{prior-enhancement} approaches do not address the fundamental need for large-scale, diverse \emph{training data} to build generalizable models.


This leaves a critical gap: a method that can generate \emph{photorealistic, multi-view consistent data} with \emph{diverse and accurate camera trajectories} to directly scale up the training of generalizable 3D vision models. 

\paragraph{Enhancing 3DGS Representations.}
Beyond the use of external priors, other works focus on improving the intrinsic properties of 3DGS. One approach leverages statistical measures to improve training and prune artifacts. For instance, Bayes' rays~\cite{DBLP:conf/cvpr/GoliRSJT24} and FisherRF~\cite{DBLP:journals/corr/abs-2311-17874} use Fisher information to quantify per-point uncertainty, guiding view selection and post-training pruning of unreliable Gaussians.
\section{Preliminary}
\label{sec:pre}
\noindent\textbf{Task definition. }
Our objective is to scale and augment 3D datasets to overcome their current limitations. This generates data with more diverse camera motions and geometric complexity, enhancing the generalization of downstream feedforward models.
%
Formally, a dataset $\mathcal{D} = \{ S_k \mid k=1, \dots, M \}$ is organized as a collection of $M$ independent scenes. Each scene $S_k$ comprises a limited set of image-pose pairs, $S_k = \{(I_{k, i}, C_{k, i}) \}_{i=1}^{N_k}$.
%
However, the set \(S_k\) provides only partial coverage of the real scene, as the geometric and photometric variations in scene \(k\) contain far more information than what is observed from the limited \(N_k\) viewpoints.
To overcome this data sparsity and unlock the full information potential of the scene, we propose to leverage reconstructed scene geometry. This reconstruction step transforms the sparse input $S_k$ into a continuous 3D representation $\mathscr{G}_k$. Hence, we can freely sample an arbitrary number of novel views, which we define as ``free-views" $\mathcal{F}_k = \{ (I^{\text{fv}}_{k, i}, C^{\text{fv}}_{k, i}) \mid i=1, \dots, N^{\text{fv}}_k \}$, achieving the desired data scaling for robust, generalizable model training.

\vspace{2pt}
\noindent\textbf{Scene reconstruction with Gaussians.}
3D Gaussians (3DGS)~\cite{kerbl20233d} achieves high-fidelity scene reconstruction by representing the scene as a set of $N$ individual 3D Gaussians $\mathscr{G}=\{g_i\}_{i=1}^{N}$. Each Gaussian $g_i$ is parameterized by the learnable properties including $\mathbf{\mu}_i \in \mathbb{R}^3$
the center of $g_i$, scaling $\mathbf{s}_i \in \mathbb{R}^3$ defining the principal axes lengths, quaternionc $\mathbf{q}_i \in \mathbb{R}^4$ and opacity $\alpha_i \in \mathbb{R}$.
\label{sec:method}
\section{Method}
The overall pipeline of FreeScale, as illustrated in Figure~\ref{fig:method}, initiates by leveraging a reconstructed 3D Gaussians derived from sparse multi-view inputs $S_k$. Our core innovation lies in a certainty-guided sampling strategy that selectively generates a large quantity of high-diversity and high-quality free-view images $\textbf{I}^\text{fv}$ and corresponding camera pose $\textbf{C}^\text{fv}$. These synthesized free-views serve two critical purposes: first, they act as powerful data augmentation for training robust downstream feedforward models (Sec. \ref{sec:jointtrain}); and second, they can be seamlessly integrated into the original reconstruction optimization loop to further enhance the quality and geometric fidelity of the initial 3D Gaussian representation (Sec. \ref{sec:gsrefine}).

\subsection{Certainty-aware Free-Views Synthesis}
To facilitate the training of robust downstream models, our primary objective is to collect a set of high-diversity and high-quality free-view images that effectively serve as data augmentation. Crucially, these desired free-views must extend beyond the support of the original training viewpoint distribution, yet still capture sufficient scene information without being dominated by large, featureless, or out-of-scene background regions. 

Since the reconstructed scene geometry $\mathscr{G}$ explicitly encode the geometry and density of the current scene, we leverage this representation to establish a sampling strategy guided by certainty. 
\change{Instead of using a fixed unit size, we discretize the scene's bounding box into a relative voxel grid of resolution $R^3$. Each voxel $v_{i}$ within this normalized grid is defined by its spatial index $i=(x, y, z)$.
We empirically set $R = 128$ to balance efficiency and accuracy; while higher resolutions in small scenes introduce redundancy, lower resolutions in large scenes yield inaccurate WIoU estimates in Eq.~\ref{eq-iou} and erroneous NMS.}

Then, The certainty of each grid $\mathcal{C}(v_i)$, is defined by accumulating the individual certainty scores of all Gaussian centers $\mathbf{\mu}_j$ that fall within its boundaries, like:
\begin{equation}
\begin{split}
    & \mathcal{C}({v_i}) = \sum_{g_j \in \mathscr{G}_i} \frac{\alpha_j}{\text{Vol}_j + \epsilon} \\
    & \text{Vol}_j \triangleq \prod_{k=1}^{3} \exp((\mathbf{s}_j)_k)
\end{split}
\label{eq-grid}
\end{equation}
where $\mathscr{G}_i = \{g_j \mid \mathbf{\mu}_j \in v_i, g_j \in \mathscr{G}\}$ denotes subset of Gaussians whose centers fall within voxel $v_i$
The certainty score highlights regions of small, opaque Gaussians, providing robust guidance for high-fidelity free-view synthesis.

\subsubsection{Virtual Viewpoints Placement}
\begin{figure}[h]
\centering
\includegraphics[width=0.48\textwidth]{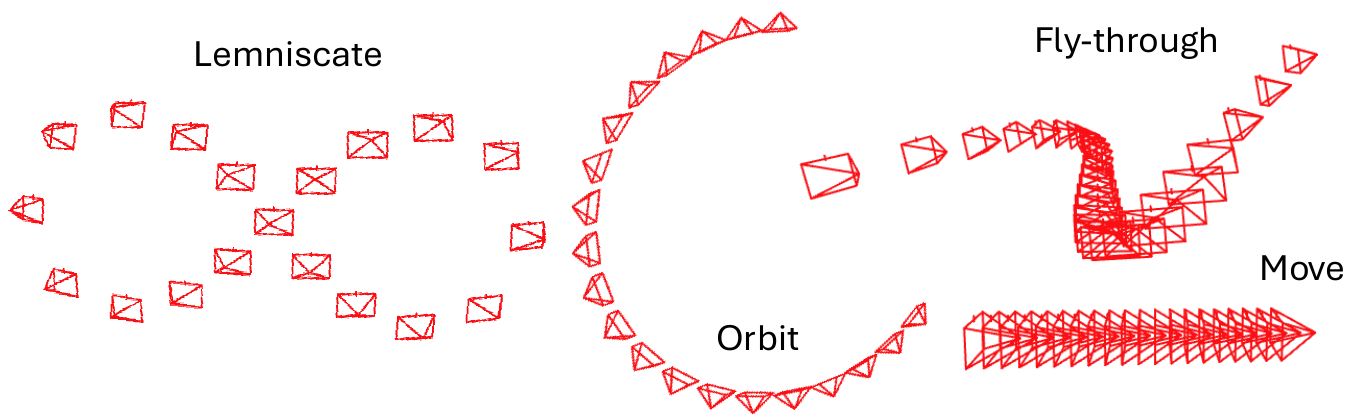}
   \vspace{-6mm}
   \caption{Showcase of predefined camera trajectory modes.}
\label{fig:traj}
\vspace{-4mm}
\end{figure}

To ensure comprehensive scene coverage and maximize viewpoint diversity, we establish ten distinct camera trajectory modes within the 3DGS $\mathscr{G}$, shown in Figure~\ref{fig:traj}. These modes include structured motions such as $\texttt{orbit}$, $\texttt{spiral}$, $\texttt{lemniscate}$, alongside various customized $\texttt{move}$ and $\texttt{fly-through}$ patterns. The generation process for candidate viewpoints $\textbf{C}^\text{fv}$ is as follows:

\begin{itemize}
    \item Anchor Selection: Trajectories are initialized from a set of anchor poses selected from the training cameras $\textbf{C}_k$ using clustering and random choice. 
    
    \item Certainty Localization: The look-at point for each generated trajectory is dynamically determined to enhance scene coverage. The look-at point of the $\texttt{move}$ and $\texttt{fly-through}$ modes adheres to the direction of the anchor camera to ensure spatial continuity and guided exploration. Conversely, for object-centric modes, the look-at point is randomly sampled from the Certainty Grid $\mathcal{C}$ within the scene's central volumetric region. This prioritization maximizes the geometric information content of the synthesized free-views by targeting well-reconstructed, high-certainty areas.
    
    \item Pose Jitter: To enhance pose diversity and mitigate potential sampling bias introduced by the pre-defined modes, a random subset of the generated poses undergoes slight rotational and translational perturbations.
\end{itemize}

\begin{table*}[t]
\centering
\caption{\textbf{Quantitative comparison of feed-forward models on viewpoint generalization.} Joint training with our FVGen data yields consistent improvements across both small and large camera motion settings. 
}
\vspace{-0.12in}
\setlength{\tabcolsep}{0.8em}
\renewcommand{\arraystretch}{1}
\resizebox{1.0\linewidth}{!}{
\begin{tabular}{lccc|ccc|ccc}
\toprule
&  \multicolumn{3}{|c|}{\textbf{In-Domain}} & \multicolumn{6}{c}{\textbf{Out-of-Domain} (Zero-shot Generalization)} \\
&  \multicolumn{3}{|c|}{\textit{DL3DV}} & \multicolumn{3}{c|}{\textit{MipNeRF360}} & \multicolumn{3}{c}{\textit{Tanks \& Temples}} \\
Method & \multicolumn{1}{|c|}{PSNR$_\uparrow$} & \multicolumn{1}{c|}{SSIM$_\uparrow$} & \multicolumn{1}{c|}{LPIPS$_\downarrow$}  & \multicolumn{1}{c|}{PSNR$_\uparrow$} & \multicolumn{1}{c|}{SSIM$_\uparrow$} & \multicolumn{1}{c|}{LPIPS$_\downarrow$} & \multicolumn{1}{c|}{PSNR$_\uparrow$} & \multicolumn{1}{c|}{SSIM$_\uparrow$} & \multicolumn{1}{c}{LPIPS$_\downarrow$}  \\
\hline \hline
\textit{\footnotesize{Small camera motion}} & \multicolumn{9}{l}{} \\

LVSM~\cite{jin2024lvsm} & \multicolumn{1}{|c}{22.20} & 0.680 & 0.216 & 15.84 & 0.285 & 0.583 & 13.07 & 0.336 & 0.674\\
LVSM w/ FreeScale & \multicolumn{1}{|c}{\textbf{24.20}} & \textbf{0.767} & \textbf{0.165} & \textbf{18.30} & \textbf{0.386} &\textbf{ 0.460} & \textbf{13.80} &\textbf{ 0.652} & \textbf{0.361}\\
\hline
\textit{\footnotesize{Large camera motion}} & \multicolumn{9}{c}{} \\
3DGS~\cite{kerbl20233d} & \multicolumn{1}{|c}{16.22} & 0.592 & 0.345  &13.47 & 0.334& 0.529& 12.12 & 0.351 & \textbf{0.569} \\
LVSM~\cite{jin2024lvsm} & \multicolumn{1}{|c}{18.75} & 0.522 & 0.352 & 13.88 & 0.293& 0.622 & 13.89 & 0.352 & 0.650\\
LVSM w/ FreeScale & \multicolumn{1}{|c}{\textbf{21.45}}  & \textbf{0.661} & \textbf{0.247} & \textbf{17.27} & \textbf{0.432} & \textbf{0.398} & \textbf{14.67} & \textbf{0.391} & 0.609\\
\bottomrule
\end{tabular}
}
\vspace{-1em}
\label{table:feedforward}
\end{table*}

\subsubsection{Virtual Viewpoints Selection with View Graph}
To ensure maximal spatial diversity, we generate a substantial pool of $N_{\text{cand}} > 2000$ candidate views per scene. However, there are plenty of redundant and poor-quality candidates. Since we focus on the high-quality viewpoints and need to keep their diversity. The intuitive idea is using image-based methods (e.g., rendering and feature matching), but they are computationally prohibitive for such a large and redundant set. Therefore, we construct a view graph to facilitate efficient view selection.


\noindent\textbf{View graph construction.}
We utilize estimated voxel certainty $\mathcal{C}(v_k)$ (as defined in Eq.~\ref{eq-grid}) to quantify view information. Each original and candidate pose serves as a node. The weighted visibility $W_{i, k}$ for a voxel $v_k$ in view $i$:
\begin{equation}
W_{i, k} = \mathcal{C}(v_k) \cdot M_{i, k}
\end{equation}
where $M_{i, k}$ is a binary visibility mask indicating whether voxel $v_k$ can be projected onto the camera plane of view $i$.
Then, the edge between any two nodes $i$ and $j$ is quantified by the Weighted Intersection-over-Union (WIoU), which measures the overlap of high-certainty information:
\begin{equation}
    \text{WIoU}(i, j) = \frac{\sum_{k=1}^{N} \min(W_{i, k} , W_{j, k})}{\sum_{k=1}^{N} \max(W_{i, k} , W_{j, k})}
\label{eq-iou}
\end{equation}
The node score $f$ for view $C_i$ is the total aggregated weighted visibility: $f(C_i) = \sum_{v_k \in \mathcal{V}} W_{i, k}$.
%
%


\subsubsection{Free-View Refinement and Rectification}
After view graph is established, we collect synthesized virtual cameras $\textbf{C}^\text{fv}$ and render images $\textbf{I}^\text{fv}$ as the initial free-view set $\mathcal{F}=(\mathbf{I}^\text{fv}, \mathbf{C}^\text{fv})$. Then, we refine  $\textbf{I}^\text{fv} \in\mathcal{F}$  to 
produce high-fidelity and photo-realistic results, and keep accurate pose placement with an iterative process of quality assessment, pose rectification, and final image enhancement.

\noindent\textbf{Image Quality Assessment and Semantic Filtering.} 
Given an established view graph, we further prune low-quality nodes with a semantic filter. 
This step is crucial since even with a small deviation of camera poses from the training cameras, the rendered images can be very blurry and therefore not useful.
We employ a combined metric for assessment: a no-reference image quality score BRISQUE~\cite{mittal2011blind} to assess visual fidelity, and a normalized depth range metric to confirm sufficient geometric content. Nodes are immediately rejected if the black pixel ratio or the normalized depth range falls below respective thresholds, indicating low quality or trivial content.

\noindent\textbf{Pose Rectification via Interpolation.}
When a candidate pose fails the image quality check, we do not discard the pose $\mathbf{C}^\text{fv}$ outright. Instead, we implement a pose rectification strategy that moves the rejected candidate closer to its nearest anchor pose. Specifically, we perform pose interpolation to gradually shift $\mathbf{C}^\text{fv}$ towards this high-quality anchor. This pose refinement effectively recovers geometrically valuable candidates that would otherwise be discarded, ensuring a robust and sufficient final pool of high-quality, high-diversity free-views $\mathcal{F}_k$.

\noindent\textbf{View-graph Guided Image Rectification.}
Although the refined free-views $\mathbf{I}^\text{fv}$ include well-reconstructed regions, underconstrained regions often still present noticeable artifacts, 
\change{falling short of full photorealism.}
To address this, we apply a \change{one step} diffusion model \change{DIFIX3D~\cite{wu2025difix3d+}} to enhance \change{the photorealism of the free-view images $\mathbf{I}^\text{fv}$.}
\change{Because the performance of diffusion enhancement is highly sensitive to the selected reference view, we use the pre-constructed view graph to guide reference selection, instead of the nearest reference based on pose distance in~\cite{wu2025difix3d+}}.
Through second-order connections, the graph captures geometric relationships that simple spatial proximity cannot, preventing the selection of ``noisy'' references that would otherwise introduce misalignment artifacts.

\begin{figure*}[h]
\centering
\includegraphics[width=1\textwidth]{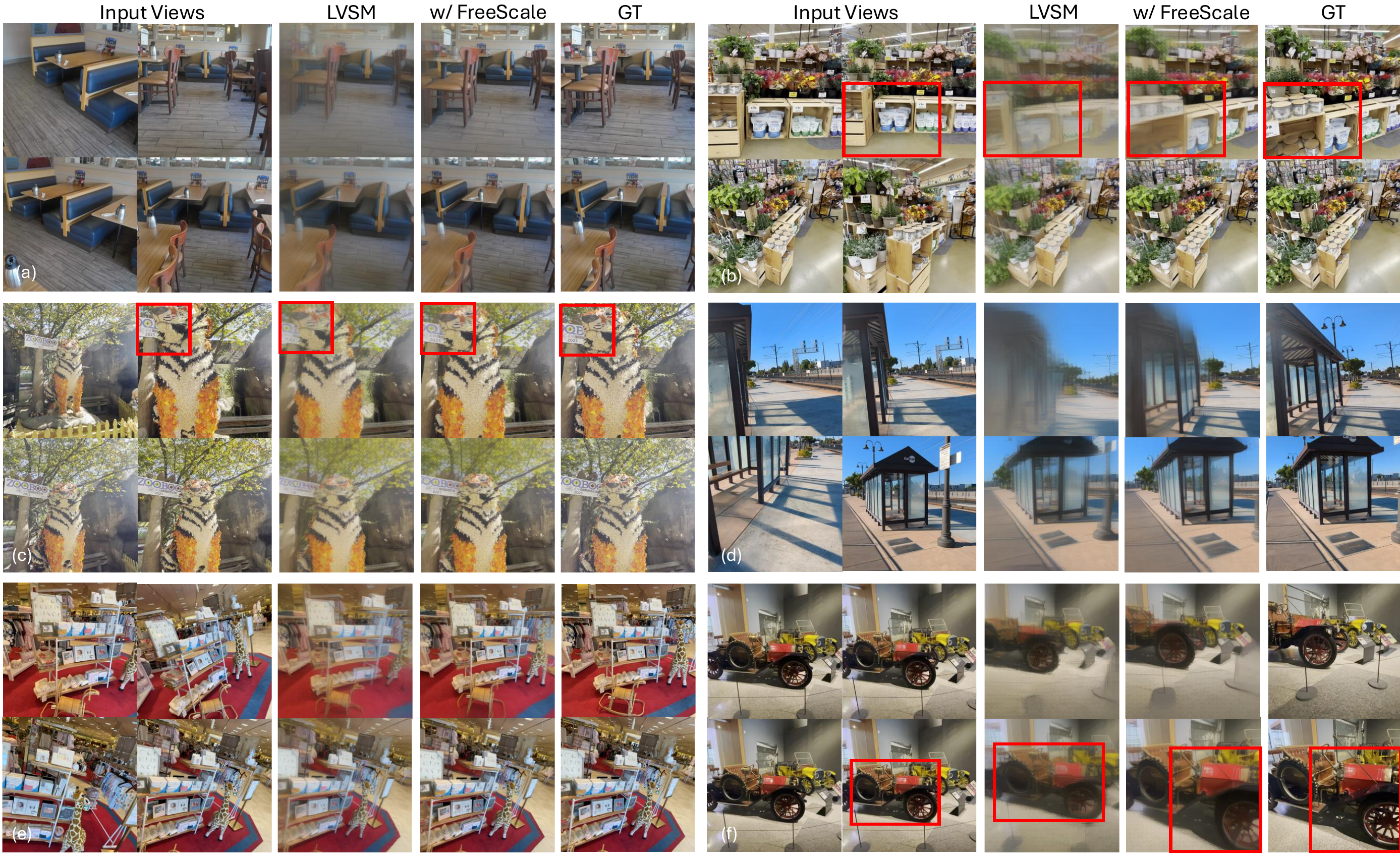}
   \vspace{-8mm}
   \caption{\textbf{Qualitative comparison for feed-forward model.}
    We use red boxes to highlight challenging regions where the target viewpoint differs significantly from the corresponding areas in the training poses.
   }
\label{fig:ff}
\vspace{-0.5em}
\end{figure*}

\subsection{Scaling Novel View Synthesis with FreeScale}
In this part, we discuss how we utilize FreeScale to scale up downstream tasks training, including feedforward models and per-scene optimization.

\subsubsection{View-graph Guided Feedforward Model Training}\label{sec:jointtrain}
Training feedforward view synthesis models~\cite{jin2024lvsm,jiang2025rayzer} requires a robust view selection strategy to construct training batches. While effective training benefits from greater camera motion diversity to ensure generalization, excessive motion can destabilize the model. Previous methods~\cite{jin2024lvsm,jiang2025rayzer} based on frame distance within the image sequence, however, confine training to this limited range of camera motion, which is suboptimal for achieving robust, generalizable view synthesis. Furthermore, the filtered free-views generated by our FreeScale do not necessarily maintain a smooth, sequential trajectory, making frame-distance sampling inappropriate.

To resolve this trade-off between motion stability and diversity, we utilize the explicit geometric and content correspondence encoded in the View Graph generated by our FreeScale to guide view selection. And we further introduce a curriculum learning mechanism to manage training stability while increasing motion diversity.

\noindent\textbf{Graph-Based View Selection.} We select input batches based on the adjacency relations of start point within the view graph, rather than relying on sequential frame indices. Our method inherently connects views by their geometric and content similarity (via WIoU), whereas frame distance only guarantees temporal proximity.

\noindent\textbf{Curriculum Learning Strategy.}
At warm-up iterations, we prioritize stability by selecting a start point that has the highest total WIoU score with its neighbors (i.e., the most information-rich or well-covered view). Input views are then randomly sampled from their high-WIoU neighbors. At the later stages, selection criteria progressively shift to favor views with lower average WIoU or greater geometric separation (as determined by lower edge weights), encouraging the model to generalize over larger camera motions and reconstruct less-constrained regions.

\subsubsection{Certainty-guided Per-Scene Reconstruction}
\label{sec:gsrefine}
Despite the high quality and increased diversity of viewpoints and appearances achieved by our generated free-views, a remaining synthetic-to-real domain gap remains. Furthermore, the explicit nature of optimization-based 3DGS often makes it highly sensitive to inconsistencies or conflicting geometric gradients in the training data, which can lead to artifacts.
To mitigate these issues, we integrate the enhanced free-views as pseudo ground truth during training. This strategy fundamentally differs from prior work~\cite{wu2025difix3d+} in how the additional views are generated and selected.
%
While~\cite{wu2025difix3d+} leverages test poses to interpolate virtual cameras, our approach is designed to operate without any knowledge of the test camera distribution.
We utilize the established view graph to automatically select free-views for training. Specially, we select the top-$K$ free-views that exhibit the lowest WIoU with the existing training cameras. These low-WIoU views represent the maximal information complement to the original dataset and are added to the training set as auxiliary targets.
The loss $\mathcal{L}_\text{FV}$ for these selected free-views is then defined as:
\begin{equation}
    \mathcal{L}_\text{FV}= \alpha^\text{fv}(||I-I^\text{fv}||_1 + ( 1-\mathcal{L}_{\text{SSIM}}(I,I^\text{fv})))
\end{equation}
where $\alpha^\text{fv}$ is a decay weight based on free-view  quality.
\section{Experiments}
\label{sec:exp}

\begin{table*}[t]
\centering
\caption{\textbf{Quantitative comparison of per-scene reconstruction on the Out-Of-Domain protocol.} Our FreeScale achieves consistent advantages in PSNR and SSIM without incurring a significant increase in computational burden.}
\label{table:perscene}
\vspace{-0.12in}  
\renewcommand{\arraystretch}{1.25}
\resizebox{1.0\linewidth}{!}{
\begin{tabular}{l|cccc|ccc|ccc}
\toprule
&  \multicolumn{4}{c|}{\textbf{DL3DV}} & \multicolumn{3}{c|}{\textbf{Nerfburster}} & \multicolumn{3}{c}{\textbf{Tanks \& Temples}} \\
Method & \multicolumn{1}{c|}{PSNR$_\uparrow$} & \multicolumn{1}{c|}{SSIM$_\uparrow$} & \multicolumn{1}{c|}{LPIPS$_\downarrow$} & \multicolumn{1}{c|}{Time (min.)$_\downarrow$} &
\multicolumn{1}{c|}{PSNR$_\uparrow$} & \multicolumn{1}{c|}{SSIM$_\uparrow$} & \multicolumn{1}{c|}{LPIPS$_\downarrow$} & 
\multicolumn{1}{c|}{PSNR$_\uparrow$} & \multicolumn{1}{c|}{SSIM$_\uparrow$} & \multicolumn{1}{c}{LPIPS$_\downarrow$} \\
\hline \hline
Nerfbusters~\cite{Nerfbusters2023} & 17.45 & 0.606 & 0.370 & -& 17.72 & 0.647 & 0.352 & - & - & - \\
DIFIX3D+~\cite{wu2025difix3d+} & 17.99 & 0.601 & 0.293 & 81.40 & 18.07 & 0.642 & 0.279 & 18.59 & 0.623 & 0.317 \\
\hline
3DGS~\cite{kerbl20233d} & 19.18 & 0.714 & 0.233 & \textbf{35.19} & 18.14 & 0.643 & 0.265 & 20.37 & 0.680 & 0.253\\
3DGS w/ DIFIX3D & 19.12 & 0.680 & \textbf{0.211} & 39.75 & 17.69  & 0.606 & 0.264 & 19.75 & 0.630 & \textbf{0.210}\\
3DGS w/ Depth & 19.07 & 0.718 & 0.227 & \underline{36.67} & 17.54 & 0.630 & 0.285 & 20.24 & 0.678 & 0.259 \\
\hline
3DGS w/ FreeScale & \textbf{19.57} & \textbf{0.723} & \underline{0.219}  & 37.22 & \textbf{18.40} & \textbf{0.648} & \textbf{0.258} & \textbf{20.66} & \textbf{0.685} & \underline{0.251} \\
\bottomrule
\end{tabular}
}
\end{table*}

\begin{figure*}[h]
\centering
\includegraphics[width=0.9\textwidth]{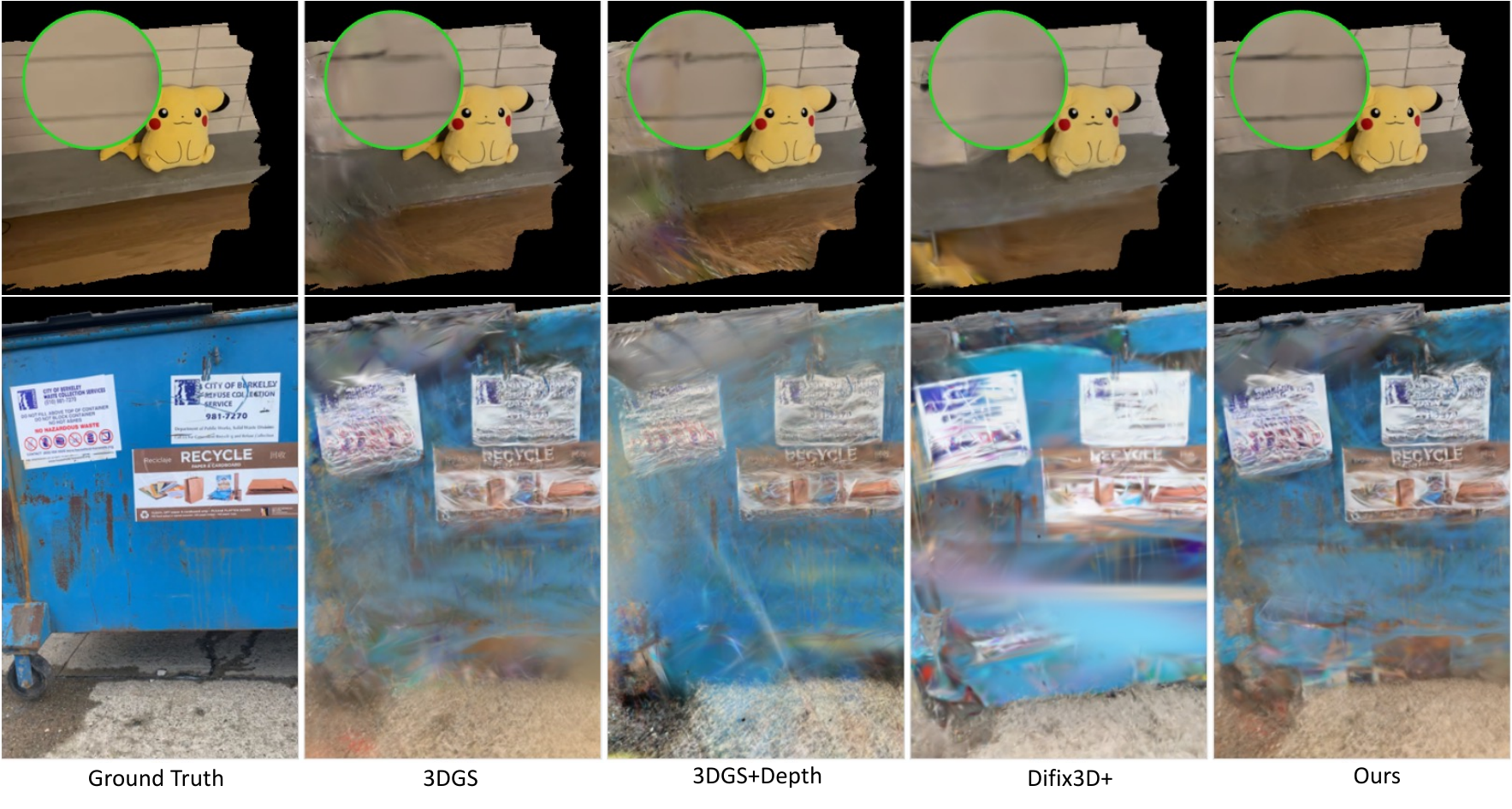}
   \vspace{-4mm}
   \caption{\textbf{Qualitative results on the Nerfbusters dataset.} The 3DGS baseline exhibits significant artifacts in unobserved areas, such as floaters and geometric noise, particularly in unobserved areas. In contrast, our method ensures high-fidelity results by sampling supplementary views from the reconstructed scene geometry.
   }
\label{fig:nerfbuster}
\vspace{-4mm}
\end{figure*}



\subsection{Improving Feed-forward NVS}

\noindent\textbf{Baselines.}
We select LVSM~\cite{jin2024lvsm} as our primary baseline, following a challenging sparse-view reconstruction paradigm, using 4 input views to predict 2 target views.

\noindent\textbf{Datasets.}
\change{We train our models on the 1,900-scene training split and evaluate on the official 110-scene benchmark of DL3DV-10K~\cite{ling2024dl3dv}.}
To ensure data integrity, we apply a pre-processing filter to discard scenes where the image count mismatches the corresponding COLMAP data or that contain corrupted image files. To assess generalization, we conduct two types of evaluations: 1) Out-of-Domain (OOD) Evaluation: To measure the generalization gains conferred by FreeScale, we conduct evaluations on extra real-world datasets MipNeRF360~\cite{barron2022mip} and Tanks \& Temples (tnt)~\cite{knapitsch2017tanks}, comprising a total of 16 scenes.
2) Viewpoint Generalization: To evaluate the model's ability to adapt to diverse camera motions, we establish two validation protocols based on frame distance. For the small camera motion setting, we sample poses, ensuring the maximum frame distance between any input and target view is no greater than 20. Conversely, for the large camera motion setting, we ensure the minimum frame distance is at least 20.

\noindent\textbf{Results.} 
Quantitative results in Table~\ref{table:feedforward} demonstrate that joint training with our FreeScale data yields consistent improvements across both small and large camera motion settings. The benefit is particularly pronounced in the challenging large camera motion scenario, where our method achieves a substantial 2.7 dB PSNR gain over the baseline.
Qualitative comparisons are presented in Figure~\ref{fig:ff}. The baseline LVSM exhibits significant difficulty with large camera motion, tending to merely replicate input views rather than generalizing to the expected novel perspective. As highlighted in the red box, the baseline's rendered viewpoint clearly regresses towards an input pose. In contrast, FreeScale provides the feed-forward model with richer priors by sampling from realistic scene geometry. As shown in Figure~\ref{fig:ff}(a), the data augmentation from our synthesized zoom-in trajectories enables the model to generate a sharper and more geometrically accurate image.

\begin{table*}[ht]
\vspace{-1.5em}
    \centering
\begin{minipage}{0.4\textwidth}
        \centering
         \caption{\textbf{Ablation of per-scene optimization on DL3DV.} ``w/ dist ref'': distance-based reference for rectification. ``w/ sparse init.'': incomplete initialization.}
         \vspace{-1em}
        \setlength{\tabcolsep}{0.5em}
        \renewcommand{\arraystretch}{1}
        \resizebox{\linewidth}{!}{
            \begin{tabular}{l c >{\columncolor{red!10}}c >{\columncolor{red!10}}c c}
            \toprule
            Methods & 3DGS & w/ dist ref & w/ sparse init. & w/ FreeScale \\
            \hline
            PSNR$_\uparrow$ & 19.18 & 17.88 & 19.51 & \textbf{19.57} \\
            SSIM$_\uparrow$ & 0.714 & 0.666 & 0.717 & \textbf{0.723} \\
            LPIPS$_\downarrow$ & 0.233 & 0.302 & 0.232 & \textbf{0.219} \\
            \bottomrule
            \end{tabular}
        }
        \label{table:rebuttal-dm}
    \end{minipage}
\vspace{-0.5em}
\hfill 
\begin{minipage}{0.58\textwidth}
    \centering
\caption{\textbf{Ablation study of different data sparsity on tnt
dataset.} This shows that our FVGen is robust and not limited to a
specific data density, providing consistent quality improvements.}
\vspace{-1em}
\setlength{\tabcolsep}{0.95em}
\renewcommand{\arraystretch}{1.1}
\resizebox{1.0\linewidth}{!}{
\begin{tabular}{l|cc|cc|cc|cc}
\toprule
&  \multicolumn{2}{c|}{10\%} & \multicolumn{2}{c|}{20\%}  & \multicolumn{2}{c|}{40\%} & \multicolumn{2}{c}{50\%}\\
Method & PSNR & SSIM & PSNR & SSIM & PSNR & SSIM & PSNR & SSIM \\
\hline
3DGS & 20.86 & 0.674 & 20.37 & 0.680  & 18.96 & 0.649 & 17.83 & 0.620 \\
FreeScale & 21.09 & 0.678 & 20.65 & 0.685 & 19.19 & 0.654 & 18.10 & 0.633 \\
\bottomrule
\end{tabular}
}
\label{table:abs-frac}
\end{minipage}

\vspace{-0.5em}
    
\end{table*}

\subsection{Enhancing Per-Scene Reconstruction}

\noindent\textbf{Baseline. }
We compare our method against 3DGS and its variants, as well as diffusion-based refinement methods like DIFIX3D~\cite{wu2025difix3d+} and Nerfbusters~\cite{Nerfbusters2023}. A critical distinction is that they load pretrained weights as a starting point for fine-tuning, whereas our model is trained from scratch. 


\noindent\textbf{Datasets. }
We conduct experiments on DL3DV, Nerfbuster, and tnt, and adopt a more challenging OOD splitting strategy. We partition each scene's data sequentially based on the camera trajectory order. Specifically, the first 70\% of frames from each DL3DV evaluation scene and the first 80\% from each tnt scene are used for training. We follow the official benchmark protocol for Nerfbuster.

\vspace{2pt}\noindent\textbf{Results. } 
Table~\ref{table:perscene} shows that FreeScale achieves consistent advantages, achieving the highest PSNR and SSIM scores.
These significant quality improvements are achieved with a negligible increase in computational cost. The runtime of FreeScale remains comparable to the baseline. This is substantially faster than DIFIX3D+~\cite{wu2025difix3d+}, which require multiple, costly diffusion inference passes, leading to significant time overhead.
The qualitative results on Nerfbuster show that FreeScale keeps more details, which can be found in Figure~\ref{fig:nerfbuster}. Our method ensures high-fidelity results by viewpoint expansion based on the reconstructed scene geometry.


\begin{table}[t]
\centering
\caption{{\textbf{Ablation study on free-view images.} ``FV" indicate generated free-view images, ``View-graph" means graph-guided joint training and certainty-guided per-scene reconstruction. }
}
\vspace{-0.12in}
\renewcommand{\arraystretch}{1.2}
\resizebox{1\linewidth}{!}{
\begin{tabular}{l|ccc|ccc}
\toprule
&  \multicolumn{3}{c|}{\textbf{Feed-forward Model}} & \multicolumn{3}{c}{\textbf{Per-Scene Optimization}} \\
Method & PSNR$_\uparrow$ & SSIM$_\uparrow$ & LPIPS$_\downarrow$ & PSNR$_\uparrow$ & SSIM$_\uparrow$ & LPIPS$_\downarrow$ \\
\hline
Baseline & 17.75 & 0.385 & 0.465 & 19.18 & 0.714 & 0.233\\
+ FV (random) & 18.68 & 0.508 & 0.342 &  19.20 & 0.715 & 0.240\\
+View-graph & 19.11 & 0.529 & 0.322 & 19.57 & 0.723 & 0.219\\
\bottomrule
\end{tabular}
}
\vspace{-0.2in}
\label{table:abs-abs}
\end{table}

\subsection{Ablation Studies and Analysis}

\noindent\textbf{Robustness to Data Sparsity. }
We evaluate the robustness of FreeScale against data sparsity from two distinct perspectives: incomplete initialization and and sparse training data. As shown in Table~\ref{table:rebuttal-dm} (w/ sparse init.), when the initial 3DGS is reconstructed using merely 5\% of the training data, FreeScale remains resilient, effectively extracting valid geometric cues to produce informative views that surpass the baseline 3DGS. Furthermore, we conduct an ablation study varying levels of overall data sparsity on tnt dataset in Table~\ref{table:abs-frac}. FreeScale consistently outperforms the baseline across all sparsity levels in both PSNR and SSIM. Together, these results demonstrate that our data engine is not overly reliant on dense initial coverage and generalizes effectively to highly sparse real-world scenarios.


\noindent\textbf{Ablation study on free-view images. }
We analyze the impact of free-view images on two key downstream tasks as shown in Table~\ref{table:abs-abs}. (1) For the feed-forward model, we conduct an ablation on the DL3DV benchmark under the large camera motion setting, training for 8K iterations. The Baseline model, trained without free-views shows poor viewpoint generalization. Simply adding free-views sampled randomly provides a substantial boost from 17.75 to 18.68 PSNR. Finally, we develop a curriculum learning strategy based on view graph information, which achieve the best performance. (2) For per-scene optimization, randomly selecting free-views yields a negligible improvement. This suggests that intuitive data augmentation can introduce conflicting or redundant information. In contrast, using the view graph to select views that provide maximal information complement achieves the best results across all metrics

\begin{figure}[h]
\vspace{-2mm}
\centering
\includegraphics[width=0.48\textwidth]{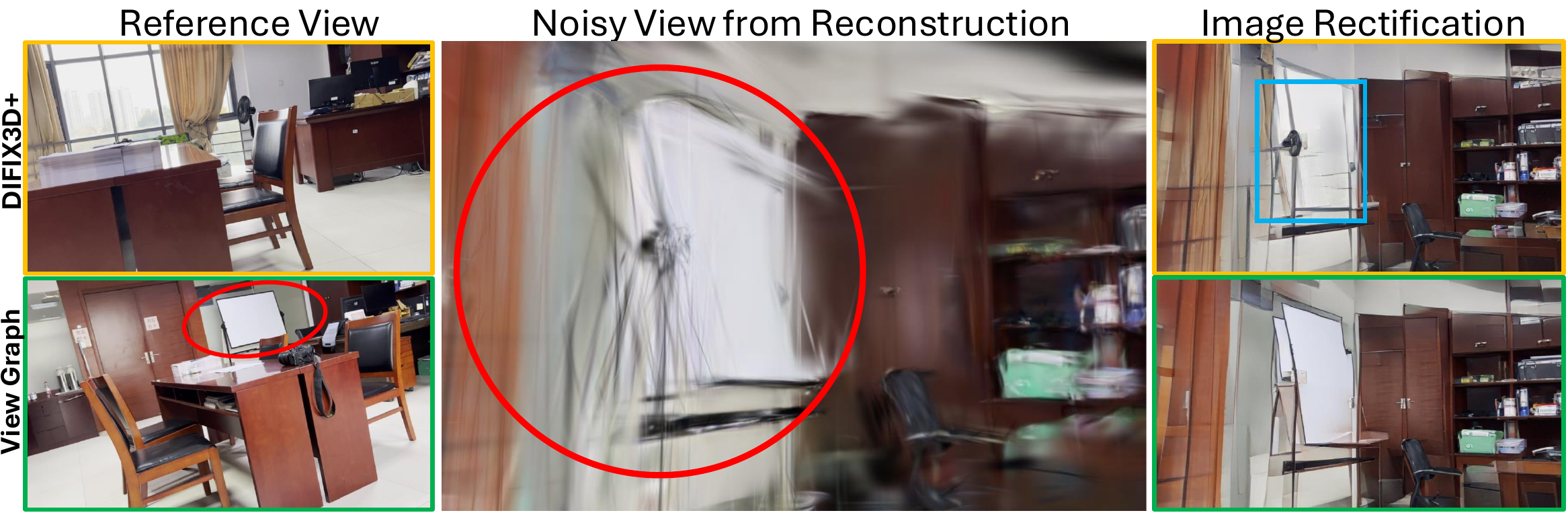}
   \vspace{-5mm}
   \caption{\textbf{Comparison of reference image selection.} Our view graph identifies the shared visible region with the noisy view (red circle), ensuring accurate image rectification.
   }
\label{fig:view_graph}
\vspace{-2mm}
\end{figure}

\vspace{2pt}

\noindent\textbf{Impact of the View Graph.}
The proposed view graph is essential for robust free-view refinement, as it provides accurate geometric correspondences between generated views and the original training cameras. Unlike previous methods that rely solely on spatial pose distance, which often fail to capture true visual overlap, our graph-based approach explicitly guarantees meaningful relationships. As shown in Figure~\ref{fig:view_graph}, the distance-based reference image selected by DIFIX3D+ lacks frustum overlap with the target view, resulting in severe rectification artifacts. Conversely, our view graph leverages shared visibility to prevent such misalignment. Quantitative results in Table~\ref{table:rebuttal-dm} further validate this: such a distance-based reference strategy (w/ dist ref) leads to noticeable performance drops, whereas our graph-guided selection consistently ensures photorealism.



\section{Conclusion}

In this work, we address the critical data bottleneck in novel view synthesis, where sparse real-world data limits generalization and synthetic data suffers from domain gaps. We introduce FreeScale, a novel data engine that transforms discrete scene data into a continuous 3D representation to generate diverse, high-fidelity free-views with accurate poses. At its core, a certainty-based view graph efficiently filters candidate viewpoints and guides image rectification, maximizing information in under-constrained regions while mitigating reconstruction artifacts. Experiments show that FreeScale provides a scalable data augmentation solution, significantly boosting downstream performance and opening a new avenue for training robust, 3D-aware models.

\paragraph{Acknowledgments} 
This work has been made possible by a Research Impact Fund project (RIF R6003-21) and a General Research Fund project (GRF 16203224) funded by the Research Grants Council (RGC) of the Hong Kong Government.
This work was partially supported by the Wallenberg AI, Autonomous Systems and Software Program (WASP) funded by the Knut and Alice Wallenberg Foundation. 
The computations resources provided by Chalmers e-Commons at Chalmers and the National Academic Infrastructure for Supercomputing in Sweden (NAISS), partially funded by the Swedish Research Council through grant agreement no. 2022-06725.

{
    \small
    \bibliographystyle{ieeenat_fullname}
    \bibliography{main}

@article{kerbl20233d,
  title={3D Gaussian splatting for real-time radiance field rendering.},
  author={Kerbl, Bernhard and Kopanas, Georgios and Leimk{\"u}hler, Thomas and Drettakis, George},
  journal={ACM Trans. Graph.},
  volume={42},
  number={4},
  pages={139--1},
  year={2023}
}

@inproceedings{jiang2025rayzer,
  title={Rayzer: A self-supervised large view synthesis model},
  author={Jiang, Hanwen and Tan, Hao and Wang, Peng and Jin, Haian and Zhao, Yue and Bi, Sai and Zhang, Kai and Luan, Fujun and Sunkavalli, Kalyan and Huang, Qixing and others},
  booktitle={Proceedings of the IEEE/CVF International Conference on Computer Vision},
  pages={4918--4929},
  year={2025}
}

@inproceedings{wu2025difix3d+,
  title={Difix3d+: Improving 3d reconstructions with single-step diffusion models},
  author={Wu, Jay Zhangjie and Zhang, Yuxuan and Turki, Haithem and Ren, Xuanchi and Gao, Jun and Shou, Mike Zheng and Fidler, Sanja and Gojcic, Zan and Ling, Huan},
  booktitle={Proceedings of the Computer Vision and Pattern Recognition Conference},
  pages={26024--26035},
  year={2025}
}

@inproceedings{mittal2011blind,
  title={Blind/referenceless image spatial quality evaluator},
  author={Mittal, Anish and Moorthy, Anush K and Bovik, Alan C},
  booktitle={2011 conference record of the forty fifth asilomar conference on signals, systems and computers (ASILOMAR)},
  pages={723--727},
  year={2011},
  organization={IEEE}
}

@inproceedings{Nerfbusters2023,
  author       = {Frederik Warburg and
                  Ethan Weber and
                  Matthew Tancik and
                  Aleksander Holynski and
                  Angjoo Kanazawa},
  title        = {Nerfbusters: Removing Ghostly Artifacts from Casually Captured NeRFs},
  booktitle    = {{IEEE/CVF} International Conference on Computer Vision},
  pages        = {18074--18084},
  year         = {2023},
}

@article{jin2024lvsm,
  title={{LVSM}: A large view synthesis model with minimal 3d inductive bias},
  author={Jin, Haian and Jiang, Hanwen and Tan, Hao and Zhang, Kai and Bi, Sai and Zhang, Tianyuan and Luan, Fujun and Snavely, Noah and Xu, Zexiang},
  journal={arXiv preprint arXiv:2410.17242},
  year={2024}
}

@article{mildenhall2021nerf,
  title={Nerf: Representing scenes as neural radiance fields for view synthesis},
  author={Mildenhall, Ben and Srinivasan, Pratul P and Tancik, Matthew and Barron, Jonathan T and Ramamoorthi, Ravi and Ng, Ren},
  journal={Communications of the ACM},
  year={2021},
}

@article{xie2024lrm,
  title={Lrm-zero: Training large reconstruction models with synthesized data},
  author={Xie, Desai and Bi, Sai and Shu, Zhixin and Zhang, Kai and Xu, Zexiang and Zhou, Yi and Pirk, S{\"o}ren and Kaufman, Arie and Sun, Xin and Tan, Hao},
  journal={Advances in Neural Information Processing Systems},
  volume={37},
  pages={53285--53316},
  year={2024}
}

@inproceedings{jiang2025megasynth,
  title={Megasynth: Scaling up 3d scene reconstruction with synthesized data},
  author={Jiang, Hanwen and Xu, Zexiang and Xie, Desai and Chen, Ziwen and Jin, Haian and Luan, Fujun and Shu, Zhixin and Zhang, Kai and Bi, Sai and Sun, Xin and others},
  booktitle={Proceedings of the Computer Vision and Pattern Recognition Conference},
  pages={16441--16452},
  year={2025}
}

@article{chen2024mvsplat360,
  title={Mvsplat360: Feed-forward 360 scene synthesis from sparse views},
  author={Chen, Yuedong and Zheng, Chuanxia and Xu, Haofei and Zhuang, Bohan and Vedaldi, Andrea and Cham, Tat-Jen and Cai, Jianfei},
  journal={Advances in Neural Information Processing Systems},
  volume={37},
  pages={107064--107086},
  year={2024}
}

@inproceedings{raistrick2024infinigen,
  title={Infinigen indoors: Photorealistic indoor scenes using procedural generation},
  author={Raistrick, Alexander and Mei, Lingjie and Kayan, Karhan and Yan, David and Zuo, Yiming and Han, Beining and Wen, Hongyu and Parakh, Meenal and Alexandropoulos, Stamatis and Lipson, Lahav and others},
  booktitle={Proceedings of the IEEE/CVF Conference on Computer Vision and Pattern Recognition},
  pages={21783--21794},
  year={2024}
}

@inproceedings{roberts2021hypersim,
  title={Hypersim: A photorealistic synthetic dataset for holistic indoor scene understanding},
  author={Roberts, Mike and Ramapuram, Jason and Ranjan, Anurag and Kumar, Atulit and Bautista, Miguel Angel and Paczan, Nathan and Webb, Russ and Susskind, Joshua M},
  booktitle={Proceedings of the IEEE/CVF international conference on computer vision},
  pages={10912--10922},
  year={2021}
}

@inproceedings{srivastava2022behavior,
  title={Behavior: Benchmark for everyday household activities in virtual, interactive, and ecological environments},
  author={Srivastava, Sanjana and Li, Chengshu and Lingelbach, Michael and Mart{\'\i}n-Mart{\'\i}n, Roberto and Xia, Fei and Vainio, Kent Elliott and Lian, Zheng and Gokmen, Cem and Buch, Shyamal and Liu, Karen and others},
  booktitle={Conference on robot learning},
  pages={477--490},
  year={2022},
  organization={PMLR}
}

@inproceedings{wang2020tartanair,
  title={Tartanair: A dataset to push the limits of visual slam},
  author={Wang, Wenshan and Zhu, Delong and Wang, Xiangwei and Hu, Yaoyu and Qiu, Yuheng and Wang, Chen and Hu, Yafei and Kapoor, Ashish and Scherer, Sebastian},
  booktitle={2020 IEEE/RSJ International Conference on Intelligent Robots and Systems (IROS)},
  pages={4909--4916},
  year={2020},
  organization={IEEE}
}

@inproceedings{nair2025scaling,
  title={Scaling Transformer-Based Novel View Synthesis with Models Token Disentanglement and Synthetic Data},
  author={Nair, Nithin Gopalakrishnan and Kaza, Srinivas and Luo, Xuan and Patel, Vishal M and Lombardi, Stephen and Park, Jungyeon},
  booktitle={Proceedings of the IEEE/CVF International Conference on Computer Vision},
  pages={28567--28576},
  year={2025}
}

@inproceedings{ling2024dl3dv,
  title={Dl3dv-10k: A large-scale scene dataset for deep learning-based 3d vision},
  author={Ling, Lu and Sheng, Yichen and Tu, Zhi and Zhao, Wentian and Xin, Cheng and Wan, Kun and Yu, Lantao and Guo, Qianyu and Yu, Zixun and Lu, Yawen and others},
  booktitle={Proceedings of the IEEE/CVF Conference on Computer Vision and Pattern Recognition},
  pages={22160--22169},
  year={2024}
}

@article{liu20243dgs,
  title={3dgs-enhancer: Enhancing unbounded 3d gaussian splatting with view-consistent 2d diffusion priors},
  author={Liu, Xi and Zhou, Chaoyi and Huang, Siyu},
  journal={Advances in Neural Information Processing Systems},
  volume={37},
  pages={133305--133327},
  year={2024}
}

@inproceedings{barron2022mip,
  title={Mip-nerf 360: Unbounded anti-aliased neural radiance fields},
  author={Barron, Jonathan T and Mildenhall, Ben and Verbin, Dor and Srinivasan, Pratul P and Hedman, Peter},
  booktitle={Proceedings of the IEEE/CVF conference on computer vision and pattern recognition},
  pages={5470--5479},
  year={2022}
}

@article{knapitsch2017tanks,
  title={Tanks and temples: Benchmarking large-scale scene reconstruction},
  author={Knapitsch, Arno and Park, Jaesik and Zhou, Qian-Yi and Koltun, Vladlen},
  journal={ACM Transactions on Graphics (ToG)},
  volume={36},
  number={4},
  pages={1--13},
  year={2017},
}

@article{DBLP:journals/tog/KnapitschPZK17,
  author       = {Arno Knapitsch and
                  Jaesik Park and
                  Qian{-}Yi Zhou and
                  Vladlen Koltun},
  title        = {Tanks and temples: benchmarking large-scale scene reconstruction},
  journal      = {{ACM} Trans. Graph.},
  volume       = {36},
  number       = {4},
  pages        = {78:1--78:13},
  year         = {2017},
}

@inproceedings{DBLP:conf/cvpr/DengLZR22,
  author       = {Kangle Deng and
                  Andrew Liu and
                  Jun{-}Yan Zhu and
                  Deva Ramanan},
  title        = {Depth-supervised NeRF: Fewer Views and Faster Training for Free},
  booktitle    = {{IEEE/CVF} Conference on Computer Vision and Pattern Recognition},
  pages        = {12872--12881},
  year         = {2022},
}

@inproceedings{DBLP:conf/cvpr/RoessleBMSN22,
  author       = {Barbara Roessle and
                  Jonathan T. Barron and
                  Ben Mildenhall and
                  Pratul P. Srinivasan and
                  Matthias Nie{\ss}ner},
  title        = {Dense Depth Priors for Neural Radiance Fields from Sparse Input Views},
  booktitle    = {{IEEE/CVF} Conference on Computer Vision and Pattern Recognition},
  pages        = {12882--12891},
  year         = {2022},
}

@inproceedings{DBLP:conf/cvpr/0007Z0ZNZ024,
  author       = {Jiahe Li and
                  Jiawei Zhang and
                  Xiao Bai and
                  Jin Zheng and
                  Xin Ning and
                  Jun Zhou and
                  Lin Gu},
  title        = {DNGaussian: Optimizing Sparse-View 3D Gaussian Radiance Fields with
                  Global-Local Depth Normalization},
  booktitle    = {{IEEE/CVF} Conference on Computer Vision and Pattern Recognition},
  pages        = {20775--20785},
  year         = {2024},
}

@inproceedings{DBLP:conf/eccv/VerverasPSDZ24,
  author       = {Evangelos Ververas and
                  Rolandos Alexandros Potamias and
                  Jifei Song and
                  Jiankang Deng and
                  Stefanos Zafeiriou},
  title        = {{SAGS:} Structure-Aware 3D Gaussian Splatting},
  booktitle    = {Computer Vision - {ECCV} 2024 - 18th European Conference},
  volume       = {15077},
  pages        = {221--238},
  year         = {2024},
}

@inproceedings{DBLP:conf/eccv/ZhangHLHZ24,
  author       = {Zheng Zhang and
                  Wenbo Hu and
                  Yixing Lao and
                  Tong He and
                  Hengshuang Zhao},
  title        = {Pixel-GS: Density Control with Pixel-Aware Gradient for 3D Gaussian
                  Splatting},
  booktitle    = {Computer Vision - {ECCV} 2024 - 18th European Conference},
  volume       = {15077},
  pages        = {326--342},
  year         = {2024},
}

@inproceedings{DBLP:conf/cvpr/0005YXX0L024,
  author       = {Tao Lu and
                  Mulin Yu and
                  Linning Xu and
                  Yuanbo Xiangli and
                  Limin Wang and
                  Dahua Lin and
                  Bo Dai},
  title        = {Scaffold-GS: Structured 3D Gaussians for View-Adaptive Rendering},
  booktitle    = {{IEEE/CVF} Conference on Computer Vision and Pattern Recognition},
  pages        = {20654--20664},
  year         = {2024},
}

@inproceedings{DBLP:conf/nips/XuMP24,
  author       = {Jiacong Xu and
                  Yiqun Mei and
                  Vishal M. Patel},
  title        = {Wild-GS: Real-Time Novel View Synthesis from Unconstrained Photo Collections},
  booktitle    = {Annual Conference
                  on Neural Information Processing Systems 2024},
  year         = {2024},
}

@article{DBLP:journals/corr/abs-2506-05280,
  author       = {Nan Wang and
                  Yuantao Chen and
                  Lixing Xiao and
                  Weiqing Xiao and
                  Bohan Li and
                  Zhaoxi Chen and
                  Chongjie Ye and
                  Shaocong Xu and
                  Saining Zhang and
                  Ziyang Yan and
                  Pierre Merriaux and
                  Lei Lei and
                  Tianfan Xue and
                  Hao Zhao},
  title        = {Unifying Appearance Codes and Bilateral Grids for Driving Scene Gaussian
                  Splatting},
  journal      = {CoRR},
  volume       = {abs/2506.05280},
  year         = {2025},
}

@inproceedings{DBLP:conf/eccv/LiLDZLT24,
  author       = {Yanyan Li and
                  Chenyu Lyu and
                  Yan Di and
                  Guangyao Zhai and
                  Gim Hee Lee and
                  Federico Tombari},
  title        = {GeoGaussian: Geometry-Aware Gaussian Splatting for Scene Rendering},
  booktitle    = {Computer Vision - {ECCV} 2024 - 18th European Conference},
  volume       = {15093},
  pages        = {441--457},
  year         = {2024},
}

@article{DBLP:journals/tog/WangWGX24,
  author       = {Yuehao Wang and
                  Chaoyi Wang and
                  Bingchen Gong and
                  Tianfan Xue},
  title        = {Bilateral Guided Radiance Field Processing},
  journal      = {{ACM} Trans. Graph.},
  volume       = {43},
  number       = {4},
  pages        = {148:1--148:13},
  year         = {2024},
}

@inproceedings{DBLP:conf/cvpr/LinLTLLLLWXYY24,
  author       = {Jiaqi Lin and
                  Zhihao Li and
                  Xiao Tang and
                  Jianzhuang Liu and
                  Shiyong Liu and
                  Jiayue Liu and
                  Yangdi Lu and
                  Xiaofei Wu and
                  Songcen Xu and
                  Youliang Yan and
                  Wenming Yang},
  title        = {VastGaussian: Vast 3D Gaussians for Large Scene Reconstruction},
  booktitle    = {Conference on Computer Vision and Pattern Recognition},
  pages        = {5166--5175},
  year         = {2024},
}

@inproceedings{DBLP:conf/nips/ChenL24,
  author       = {Yu Chen and
                  Gim Hee Lee},
  title        = {{DOGS:} Distributed-Oriented Gaussian Splatting for Large-Scale 3D
                  Reconstruction Via Gaussian Consensus},
  booktitle    = {Annual Conference
                  on Neural Information Processing Systems 2024},
  year         = {2024},
}

@inproceedings{DBLP:conf/iccv/WarburgWTHK23,
  author       = {Frederik Warburg and
                  Ethan Weber and
                  Matthew Tancik and
                  Aleksander Holynski and
                  Angjoo Kanazawa},
  title        = {Nerfbusters: Removing Ghostly Artifacts from Casually Captured NeRFs},
  booktitle    = {{IEEE/CVF} International Conference on Computer Vision},
  pages        = {18074--18084},
  year         = {2023},
}

@inproceedings{DBLP:conf/cvpr/GoliRSJT24,
  author       = {Lily Goli and
                  Cody Reading and
                  Silvia Sell{\'{a}}n and
                  Alec Jacobson and
                  Andrea Tagliasacchi},
  title        = {Bayes' Rays: Uncertainty Quantification for Neural Radiance Fields},
  booktitle    = {{IEEE/CVF} Conference on Computer Vision and Pattern Recognition},
  pages        = {20061--20070},
  year         = {2024},
}

@inproceedings{DBLP:conf/eccv/ChenXZZPGCC24,
  author       = {Yuedong Chen and
                  Haofei Xu and
                  Chuanxia Zheng and
                  Bohan Zhuang and
                  Marc Pollefeys and
                  Andreas Geiger and
                  Tat{-}Jen Cham and
                  Jianfei Cai},
  title        = {MVSplat: Efficient 3D Gaussian Splatting from Sparse Multi-view Images},
  booktitle    = {18th European Conference},
  volume       = {15079},
  pages        = {370--386},
  year         = {2024},
}

@inproceedings{DBLP:conf/eccv/LiuWHSYZCLL24,
  author       = {Tianqi Liu and
                  Guangcong Wang and
                  Shoukang Hu and
                  Liao Shen and
                  Xinyi Ye and
                  Yuhang Zang and
                  Zhiguo Cao and
                  Wei Li and
                  Ziwei Liu},
  title        = {MVSGaussian: Fast Generalizable Gaussian Splatting Reconstruction
                  from Multi-View Stereo},
  booktitle    = {Computer Vision - {ECCV} 2024 - 18th European Conference},
  volume       = {15076},
  pages        = {37--53},
  year         = {2024},
}

@inproceedings{DBLP:conf/cvpr/CharatanLTS24,
  author       = {David Charatan and
                  Sizhe Lester Li and
                  Andrea Tagliasacchi and
                  Vincent Sitzmann},
  title        = {PixelSplat: 3D Gaussian Splats from Image Pairs for Scalable Generalizable
                  3D Reconstruction},
  booktitle    = {{IEEE/CVF} Conference on Computer Vision and Pattern Recognition},
  pages        = {19457--19467},
  year         = {2024},
}

@inproceedings{DBLP:conf/eccv/ZhangBTXZSX24,
  author       = {Kai Zhang and
                  Sai Bi and
                  Hao Tan and
                  Yuanbo Xiangli and
                  Nanxuan Zhao and
                  Kalyan Sunkavalli and
                  Zexiang Xu},
  title        = {{GS-LRM:} Large Reconstruction Model for 3D Gaussian Splatting},
  booktitle    = {Computer Vision - {ECCV} 2024 - 18th European Conference},
  volume       = {15080},
  pages        = {1--19},
  year         = {2024},
}

@inproceedings{DBLP:conf/cvpr/XuPWBB0P25,
  author       = {Haofei Xu and
                  Songyou Peng and
                  Fangjinhua Wang and
                  Hermann Blum and
                  Daniel Barath and
                  Andreas Geiger and
                  Marc Pollefeys},
  title        = {DepthSplat: Connecting Gaussian Splatting and Depth},
  booktitle    = {{IEEE/CVF} Conference on Computer Vision and Pattern Recognition},
  pages        = {16453--16463},
  year         = {2025},
}

@inproceedings{DBLP:conf/iclr/YeLXLP0P25,
  author       = {Botao Ye and
                  Sifei Liu and
                  Haofei Xu and
                  Xueting Li and
                  Marc Pollefeys and
                  Ming{-}Hsuan Yang and
                  Songyou Peng},
  title        = {No Pose, No Problem: Surprisingly Simple 3D Gaussian Splats from Sparse
                  Unposed Images},
  booktitle    = {The Thirteenth International Conference on Learning Representations,},
  year         = {2025},
}

@article{DBLP:journals/corr/abs-2410-12781,
  author       = {Ziwen Chen and
                  Hao Tan and
                  Kai Zhang and
                  Sai Bi and
                  Fujun Luan and
                  Yicong Hong and
                  Fuxin Li and
                  Zexiang Xu},
  title        = {Long-LRM: Long-sequence Large Reconstruction Model for Wide-coverage
                  Gaussian Splats},
  journal      = {CoRR},
  volume       = {abs/2410.12781},
  year         = {2024},
  eprinttype    = {arXiv},
  eprint       = {2410.12781},
}

@article{DBLP:journals/corr/abs-2505-23716,
  author       = {Lihan Jiang and
                  Yucheng Mao and
                  Linning Xu and
                  Tao Lu and
                  Kerui Ren and
                  Yichen Jin and
                  Xudong Xu and
                  Mulin Yu and
                  Jiangmiao Pang and
                  Feng Zhao and
                  Dahua Lin and
                  Bo Dai},
  title        = {AnySplat: Feed-forward 3D Gaussian Splatting from Unconstrained Views},
  journal      = {CoRR},
  volume       = {abs/2505.23716},
  year         = {2025},
}

@inproceedings{DBLP:conf/cvpr/Wang0CCR24,
  author       = {Shuzhe Wang and
                  Vincent Leroy and
                  Yohann Cabon and
                  Boris Chidlovskii and
                  J{\'{e}}r{\^{o}}me Revaud},
  title        = {DUSt3R: Geometric 3D Vision Made Easy},
  booktitle    = {{IEEE/CVF} Conference on Computer Vision and Pattern Recognition},
  pages        = {20697--20709},
  year         = {2024},
}

@inproceedings{DBLP:conf/cvpr/WangCKV0N25,
  author       = {Jianyuan Wang and
                  Minghao Chen and
                  Nikita Karaev and
                  Andrea Vedaldi and
                  Christian Rupprecht and
                  David Novotn{\'{y}}},
  title        = {{VGGT:} Visual Geometry Grounded Transformer},
  booktitle    = {{IEEE/CVF} Conference on Computer Vision and Pattern Recognition},
  pages        = {5294--5306},
  year         = {2025},
}

@article{DBLP:journals/corr/abs-2311-15127,
  author       = {Andreas Blattmann and
                  Tim Dockhorn and
                  Sumith Kulal and
                  Daniel Mendelevitch and
                  Maciej Kilian and
                  Dominik Lorenz and
                  Yam Levi and
                  Zion English and
                  Vikram Voleti and
                  Adam Letts and
                  Varun Jampani and
                  Robin Rombach},
  title        = {Stable Video Diffusion: Scaling Latent Video Diffusion Models to Large
                  Datasets},
  journal      = {CoRR},
  volume       = {abs/2311.15127},
  year         = {2023},
}

@inproceedings{DBLP:conf/cvpr/WuMHPGWSVBPH24,
  author       = {Rundi Wu and
                  Ben Mildenhall and
                  Philipp Henzler and
                  Keunhong Park and
                  Ruiqi Gao and
                  Daniel Watson and
                  Pratul P. Srinivasan and
                  Dor Verbin and
                  Jonathan T. Barron and
                  Ben Poole and
                  Aleksander Holynski},
  title        = {ReconFusion: 3D Reconstruction with Diffusion Priors},
  booktitle    = {{IEEE/CVF} Conference on Computer Vision and Pattern Recognition},
  pages        = {21551--21561},
  year         = {2024},
}

@article{DBLP:journals/corr/abs-2311-17874,
  author       = {Wen Jiang and
                  Boshu Lei and
                  Kostas Daniilidis},
  title        = {FisherRF: Active View Selection and Uncertainty Quantification for
                  Radiance Fields using Fisher Information},
  journal      = {CoRR},
  volume       = {abs/2311.17874},
  year         = {2023},
}

@inproceedings{DBLP:conf/cvpr/WangWGSZBMSF21,
  author       = {Qianqian Wang and
                  Zhicheng Wang and
                  Kyle Genova and
                  Pratul P. Srinivasan and
                  Howard Zhou and
                  Jonathan T. Barron and
                  Ricardo Martin{-}Brualla and
                  Noah Snavely and
                  Thomas A. Funkhouser},
  title        = {IBRNet: Learning Multi-View Image-Based Rendering},
  booktitle    = {{IEEE} Conference on Computer Vision and Pattern Recognition},
  pages        = {4690--4699},
  year         = {2021},
}

@inproceedings{DBLP:conf/cvpr/ChenL23,
  author       = {Yu Chen and
                  Gim Hee Lee},
  title        = {{DBARF:} Deep Bundle-Adjusting Generalizable Neural Radiance Fields},
  booktitle    = {{IEEE/CVF} Conference on Computer Vision and Pattern Recognition},
  pages        = {24--34},
  year         = {2023},
}

@book{hess2013blender,
  title={Blender foundations: The essential guide to learning blender 2.5},
  author={Hess, Roland},
  year={2013},
  publisher={Routledge}
}

@inproceedings{DBLP:conf/cvpr/SchonbergerF16,
  author       = {Johannes L. Sch{\"{o}}nberger and
                  Jan{-}Michael Frahm},
  title        = {Structure-from-Motion Revisited},
  booktitle    = {{IEEE} Conference on Computer Vision and Pattern Recognition},
  pages        = {4104--4113},
  year         = {2016},
}

@inproceedings{DBLP:conf/nips/KheradmandRSSTI24,
  author       = {Shakiba Kheradmand and
                  Daniel Rebain and
                  Gopal Sharma and
                  Weiwei Sun and
                  Yang{-}Che Tseng and
                  Hossam Isack and
                  Abhishek Kar and
                  Andrea Tagliasacchi and
                  Kwang Moo Yi},
  title        = {3D Gaussian Splatting as Markov Chain Monte Carlo},
  booktitle    = {Advances in Neural Information Processing Systems},
  year         = {2024},
}

@inproceedings{DBLP:conf/nips/KulhanekPKPS24,
  author       = {Jonas Kulhanek and
                  Songyou Peng and
                  Zuzana Kukelova and
                  Marc Pollefeys and
                  Torsten Sattler},
  title        = {WildGaussians: 3D Gaussian Splatting In the Wild},
  booktitle    = {Advances in Neural Information Processing Systems},
  year         = {2024},
}

@inproceedings{DBLP:conf/eccv/MorgensternBHE24,
  author       = {Wieland Morgenstern and
                  Florian Barthel and
                  Anna Hilsmann and
                  Peter Eisert},
  title        = {Compact 3D Scene Representation via Self-Organizing Gaussian Grids},
  booktitle    = {Computer Vision - {ECCV} 2024 - 18th European Conference},
  volume       = {15143},
  pages        = {18--34},
  year         = {2024},
}
}

\clearpage
\setcounter{page}{1}
\maketitlesupplementary
This supplementary material consists of three parts: technical details of the experimental setup (Sec. \ref{sec:supp-detail}), additional ablation studies on free-views generation (Sec. \ref{sec:supp_ablation}), and additional qualitative results (Sec. \ref{sec:supp_results}), including out-of-domain results and a discussion about limitations (Sec. \ref{sec:supp_limitation}).

\section{Implement Details}
\label{sec:supp-detail}
\subsection{Certainty-aware Free-View Synthesis}

\paragraph{Virtual Viewpoints Placement.}
We first generate virtual viewpoints trajectories with 10 predefined modes, including: \textbf{geometric paths}: (1) orbit, (2) spiral, (3) lemniscate; (4) \textbf{interpolation}; and \textbf{cinematic movements}: (5) move up, (6) move down, (7) move left, (8) move right, (9) dollyzoom in, (10) dollyzoom out.
For each mode, we select $N_{traj}=10$ anchor poses via K-Means clustering or Farthest Point Sampling (FPS) on the training views.
These anchor poses are randomly perturbed with position noise sampled from $\mathcal{N}(0, \sigma_{pos})$ where $\sigma_{pos} \in [0, 0.1]$ and rotation jitter within $\pm 20^{\circ}$.
We generate sequences of $L=20$ frames per trajectory, resulting in a dense candidate pool. To enhance viewpoint diversity, we apply random perturbations to the initial poses, with position noise sampled from $\mathcal{N}(0, \sigma_{pos})$ where $\sigma_{pos} \in [0, 0.5]$ and rotation jitter within $\pm 30^{\circ}$. Our candidate pool has more than 2,000 candidate views per scene.

\paragraph{Virtual Viewpoints Selection.}
To eliminate invalid views (e.g., occluded or unbounded regions), we first perform rigorous spatial feasibility checks on all candidate poses, immediately rejecting those that violate geometric constraints. This involves rejecting poses falling outside the scene's established bounding box or those situated inside known structures. Only the remaining feasible poses are considered for node status.
Then, we perform Non-Maximum Suppression (NMS) on the candidate poses based on the established view graph. The score of view $f(C_i)$ quantifies its information gain. Candidate poses are first sorted in descending order of $f(C_i)$. We initialize the selected set $\mathcal{F}_{selected}$ with all poses from the training set. A candidate pose $i$ is accepted and added to $\mathcal{F}_{selected}$ if its W-IoU with all poses $j \in \mathcal{F}_{selected}$ remains below the threshold of $0.7$. This filtering process continues until $K=500$ non-redundant candidates are successfully selected, ensuring both high certainty and diversity free-views.

\paragraph{Free-View Refinement and Rectification.}
During rendering, we enforce quality assurance using the BRISQUE metric $< 0.5$. And a depth percentile range validity score calculated on the central $70\%$ crop, must be greater than $0.1$.
If a rendered view fails these checks, we trigger a pose rectification mechanism: the camera is iteratively shifted towards the nearest training view with decreasing step distances $\{0.7, 0.5, 0.3\}$. Finally, we apply stepwise filtering based on the BRISQUE quality metric to select the target set of $\approx 100$ high-quality free-view frames, retaining all candidates if the target quota is not fulfilled.

\begin{table*}[t]
\centering
\renewcommand{\arraystretch}{1.2}
\caption{{\textbf{Ablation study on Free-View generation.} Our certainty-aware generation relies fundamentally on the certainty grid and the established view graph. \textbf{Without the view graph}, selecting the top-500 candidates solely by certainty score results in high redundancy and fails to provide valuable guidance for feed-forward model training. \textbf{Without the certainty grid}, we must resort to calculating inter-view correspondence only via position and rotation distance, which is both inaccurate and computationally inefficient.}
}
\vspace{-0.12in}
\resizebox{1\linewidth}{!}{
\begin{tabular}{lll|cccc|ccc}
\toprule
&  &&\multicolumn{4}{c|}{\textbf{FreeView Statistics}} & \multicolumn{3}{c}{\textbf{Feed-Forward Model}} \\
Method & Certainty Grid & View Graph & \#Image& Per-scene \#Image & BRISQUE$_\downarrow$ & Avg. Time (s)$_\downarrow$& PSNR$_\uparrow$ & SSIM$_\uparrow$ & LPIPS$_\downarrow$ \\
\hline
FVGen & \checkmark &  \checkmark & 145,528 & 75 & 0.36 & \textbf{225.64} & \textbf{19.11} & \textbf{0.529} &\textbf{0.322} \\
 & \checkmark & -  & 164,874 & 91 & 0.38 & 608.71 & 17.63 & 0.480 & 0.397 \\
 & - &  \checkmark &  166,282 & 109 & 0.36 & 727.93 & 17.86 & 0.491  & 0.354\\
Baseline & - & - & -&-&-&-& 17.75& 0.385 & 0.465 \\
\bottomrule
\end{tabular}
}
\vspace{-0.1in}
\label{table:supp-abs}
\end{table*}

\begin{figure*}[t]
\centering
\includegraphics[width=1\textwidth]{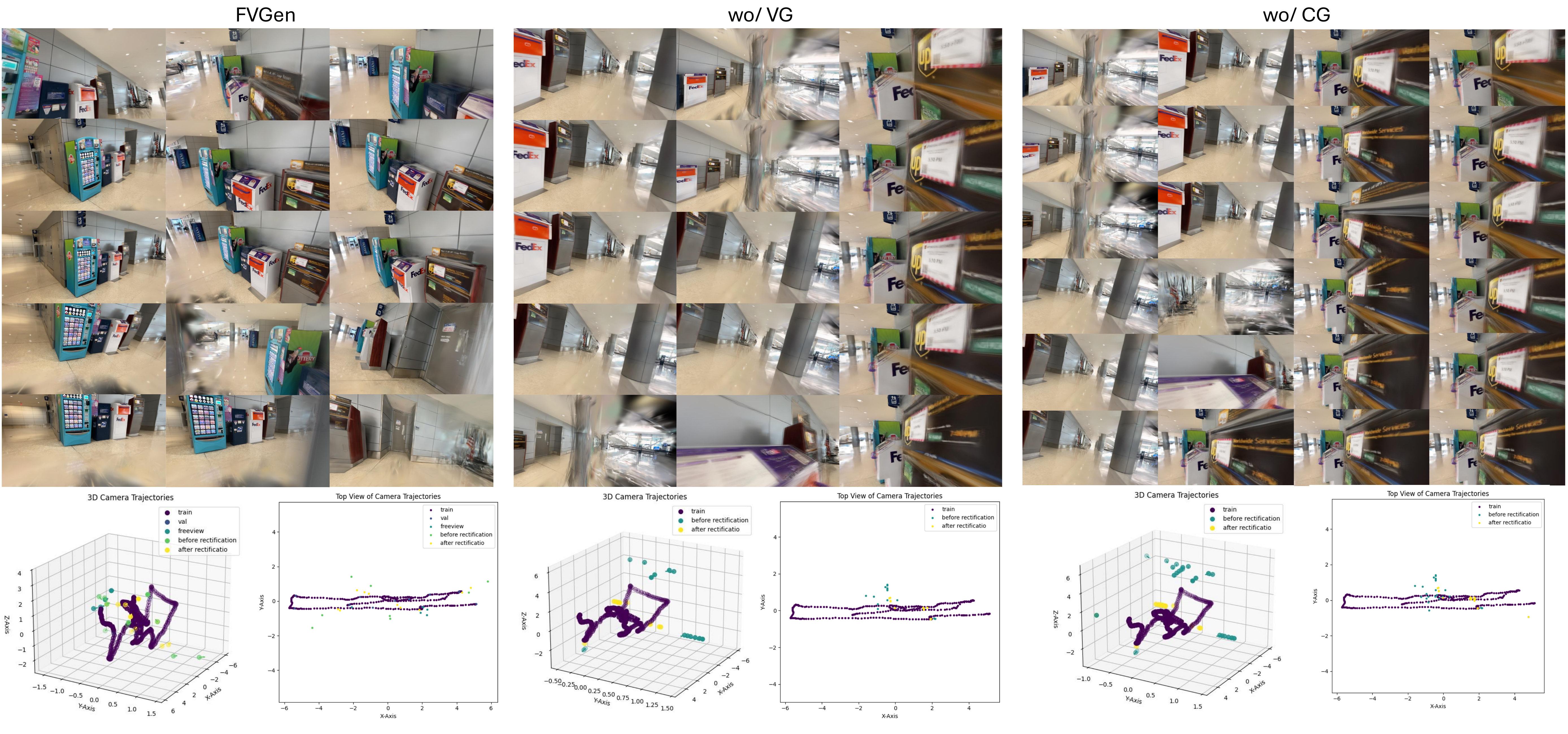}
   \vspace{-1em}
   \caption{\textbf{Showcase of different freeview generation.} Our FVGen maximally captures under-constrained geometry while being minimally contaminated by reconstruction
artifacts.
   }
\label{fig:supp-sample_examples}
\vspace{-1.5em}
\end{figure*}

\subsection{Per-Scene Reconstruction} 

\paragraph{Baseline Training.} Our 3DGS training pipeline follows the standard steps of~\cite{kerbl20233d}. We use sparse points from COLMAP~\cite{DBLP:conf/cvpr/SchonbergerF16} for initialization. The initial opacity is 0.5. We adopt the densification strategy of MCMC-3DGS~\cite{DBLP:conf/nips/KheradmandRSSTI24} for better scene representation and compression. For all datasets, we train 3DGS for 30,000 iterations; the densification step starts from 500 and ends at 25,000. We densify each scene for every 500 iterations during training and densify at most 1,500,000 3DGS primitives. To adapt to the appearance changes, we follow WildGaussians~\cite{DBLP:conf/nips/KulhanekPKPS24}, which applies an appearance feature with dim 32 per 3DGS and trains a shallow MLP as the appearance decoder. For further acceleration, we adopt tiny-cuda-nn\footnote{\url{https://github.com/NVlabs/tiny-cuda-nn}} as the shallow MLP implementation. The depth and width of the shallow MLP are, respectively, 2 and 64. Considering the disk storage for 3DGS can be large when training thousands of scenes, we
compress 3DGS using SOGS~\cite{DBLP:conf/eccv/MorgensternBHE24} to reduce the size. For all baselines (3DGS~\cite{kerbl20233d} and DIFIX3D+~\cite{wu2025difix3d+}) and our method, we adopt gsplat\footnote{\url{https://github.com/nerfstudio-project/gsplat}} as the CUDA rasterization kernel.

\paragraph{3DGS Training with Freeview Enhancement.}
Unlike existing extrapolation methods, we train 3D Gaussian Splatting (3DGS) from scratch using augmented scene data, including the generated Freeview images. We adopt an iterative pseudo-labeling strategy: for every 3k iterations, we non-recurrently select the top-5 Freeview images that exhibit the lowest W-IoU with the existing training cameras. These selected images are incorporated into the training set as pseudo-GT until all Freeview images have been added. The loss weight for each incorporated pseudo-GT is assigned a decaying factor $\alpha^{\text{fv}} \in [0.3, 0.5]$ based on its corresponding BRISQUE quality metric.

\subsection{Scaling Up LVSM}
We train LVSM for 20,000 iterations on 1,900 scenes from the DL3DV dataset, following the established setup in \cite{jin2024lvsm}. The training utilizes 4 A40 GPUs, with a batch size of 24 per GPU. For optimization, we employ a cosine learning rate schedule that peaks at $4 \times 10^{-4}$ after a 3,000-iteration warmup period.

We set the standard frame distance between input and target views to $[15, 40]$; this distance range is also applied when selecting neighboring nodes in our view graph. To stabilize early training, we implement a curriculum learning strategy during the warm-up phase: the frame distance is gradually annealed from a narrow range of $[10, 20]$ to the standard $[15, 40]$. In addition, input and target views are selected based on the view graph with a probability $50\%$ throughout the training process.

\vspace{-4pt}
\section{Additional Ablation Studies}
\vspace{-4pt}
\label{sec:supp_ablation}
In this part, we conduct more ablation studies on free-view generation and show more cases about reference image selection mentioned in the \textcolor{blue}{main body Sec.5.3}.

\vspace{-4pt}
\subsection{Ablation on Free-View Generation}
\vspace{-4pt}
The primary objective of FVGen is to collect a set of high-diversity and high-quality free-view images. This is achieved by utilizing a certainty grid to score the information value of candidate viewpoints. Crucially, we employ the established view graph to quantify inter-view correspondence, enabling the efficient filtering of redundant poses while simultaneously preserving viewpoint diversity. All ablation experiments are conducted on the same subset of DL3DV scenes. We report the feed-forward model results in Table~\ref{table:supp-abs}, using identical settings to the results presented in \textcolor{blue}{the main body Table 3}. We also showcase the generated freeviews (before image rectification) in Figure~\ref{fig:supp-sample_examples}.

\noindent\textbf{Without View Graph.}
When the view graph is omitted, the selection process relies solely on the certainty score to choose the top-500 candidates (as detailed in the \textcolor{blue}{main body, Sec. 4.1.2}). This reliance leads to high redundancy among the selected views, resulting in insufficient scene coverage, as illustrated in Figure~\ref{fig:supp-sample_examples}.

Moreover, the absence of inter-view correspondence necessitates the random integration of these generated free views into the feed-forward model training. This approach introduces significant view shifts (i.e., no overlap between input and target views), causing training instability. As shown in Table~\ref{table:supp-abs}, while this method yields more images, it results in inferior overall image quality (higher BRISQUE) and provides poor performance gains for the downstream model, where PSNR drops from $17.75$ to $17.63$ dB.

\vspace{2pt}
\noindent\textbf{Without Certainty Grid. }
As our view graph construction strongly relies on the certainty grid, its absence necessitates an alternative for establishing inter-view correspondence. We resort to calculating the combined position distance and the angular distance between quaternions. However, this approach presents several drawbacks. 1) Setting a suitable threshold for the combined distance is non-trivial; we empirically select $3.5$ to retain approximately 500 candidates.  2) The lack of certainty-based viewpoint scoring requires computing all pairwise distances between candidates for view selection, leading to excessive generation time (Table~\ref{table:supp-abs}). 3) The correspondence based solely on the position and rotation distance is inherently inaccurate, as small combined distances do not guarantee sufficient common visibility when camera rotations differ significantly, a phenomenon illustrated in Figure~\ref{fig:supp-refimage}. Consequently, the resulting inter-view correspondence leads to suboptimal performance in the downstream feed-forward model in Table~\ref{table:supp-abs}.

\begin{table*}[ht]
\centering
    \centering
    \setlength{\tabcolsep}{0.95em}
    \renewcommand{\arraystretch}{0.79}
    \resizebox{\linewidth}{!}{
        \begin{tabular}{l|ccc|ccc}
        \toprule
        Method & PSNR$_\uparrow$ & SSIM$_\uparrow$ & LPIPS$_\downarrow$ & PSNR$_\uparrow$ & SSIM$_\uparrow$ & LPIPS$_\downarrow$ \\
        \hline
        & \multicolumn{3}{c|}{\textit{\footnotesize{Small camera motion}}} & \multicolumn{3}{c}{\textit{\footnotesize{Large camera motion}}} \\
        LVSM & 22.20 & 0.680 & 0.216 & 18.75 & 0.522 & 0.352 \\
        \rowcolor{red!15}wo/ Diffusion & 23.41 & 0.736 & 0.185 & 20.89 & 0.634 & 0.268\\
        w/ FreeScale & \textbf{24.20} & \textbf{0.767} & \textbf{0.165} & \textbf{21.45} & \textbf{0.661} & \textbf{0.247}\\
        \bottomrule
        \end{tabular}
    }
    \caption{\textbf{Ablation isolating the impact of diffusion-based image rectification.} Comparing the LVSM baseline to our method without diffusion (\textit{wo/ Diffusion}) demonstrates that the primary performance boost stems directly from the expanded viewpoint diversity and geometric coverage. Integrating the diffusion prior (FreeScale) resolves remaining artifacts for optimal fidelity.}
    \label{table:rebuttal-nodm-lvsm}
\end{table*} 

\begin{figure*}[hbtp]
\centering
\includegraphics[width=1\textwidth]{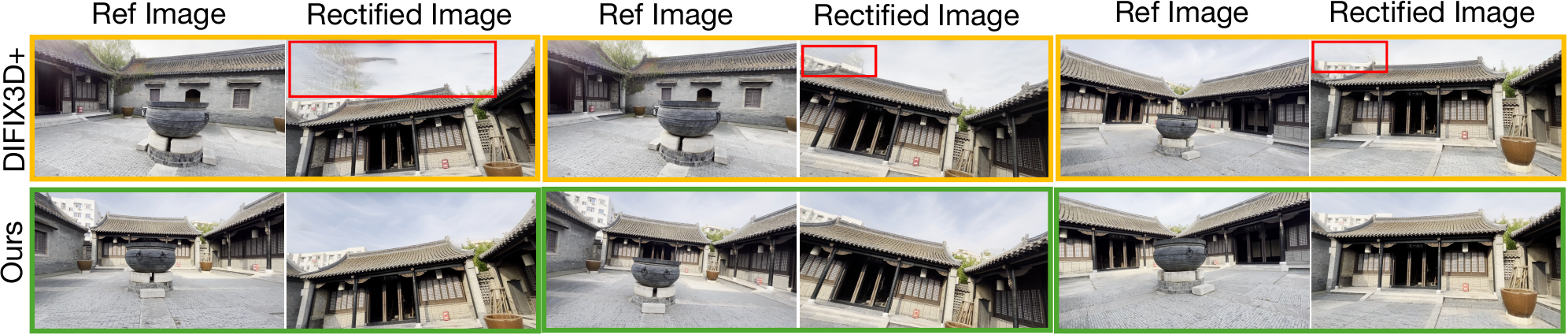}
\caption{\textbf{Consistent showcases of view graph impact.} Compared to DIFIX3D+’s distance-based reference selection strategy, our view graph provides better overlap and higher free-view consistency for reference. The red bounding boxes delineate artifacts introduced by inaccurate reference images during the image rectification stage.
}
\label{fig:appendix-inconsistency}
\end{figure*}

\subsection{Effectiveness of Free-Views without Diffusion}
To explicitly isolate the performance gain provided by generating novel viewpoints from reconstructed scene geometry, we ablate the diffusion-based image rectification. While applying 2D diffusion independently to each frame inevitably introduces minor multi-view inconsistencies (e.g., high-frequency flickering), we observe that downstream feed-forward models are remarkably robust to this issue. Because these models inherently aggregate features across multiple views, they effectively learn the underlying 3D scene geometry while filtering out inconsistent generative artifacts as noise. To explicitly isolate the contribution of the free-view images, we conduct an ablation study in Table~\ref{table:rebuttal-nodm-lvsm}. Remarkably, even without the diffusion-based refinement (\textit{w/o Diffusion}), our method still significantly outperforms the LVSM baseline. This improvement is particularly pronounced in large camera motion scenarios, yielding a substantial +2.14 dB PSNR gain. These results definitively confirm that the primary performance boost stems from the expanded viewpoint coverage and spatial diversity provided by our certainty-guided sampling, rather than solely relying on the generative prior of the diffusion model.

\subsection{More Analysis on Reference Images Selection}
As discussed in \textcolor{blue}{the main body Sec. 5.3 and Figure 6}, we provide further visualization analysis of our reference image selection strategy based on inter-view correspondence. Unlike methods such as DIFIX3D~\cite{wu2025difix3d+}, which rely on calculating a combined metric of position and rotation distance to establish correspondence, our certainty-aware view graph (VG) yields superior reference images that share a more precise common visible area with the sampled noisy view. In Figure~\ref{fig:supp-refimage}, the red circles highlight this shared visible region, while the blue bounding boxes delineate artifacts introduced by inaccurate reference images during the image rectification stage. We also show a consistent example in Figure~\ref{fig:appendix-inconsistency}. Compared to DIFIX3D+’s distance-based strategy, our view-graph-based reference selection provides better overlap and higher free-view consistency.

\subsection{Ablation on Camera Trajectory Modes}
\begin{figure*}[htbp]
\centering
\includegraphics[width=1\textwidth]{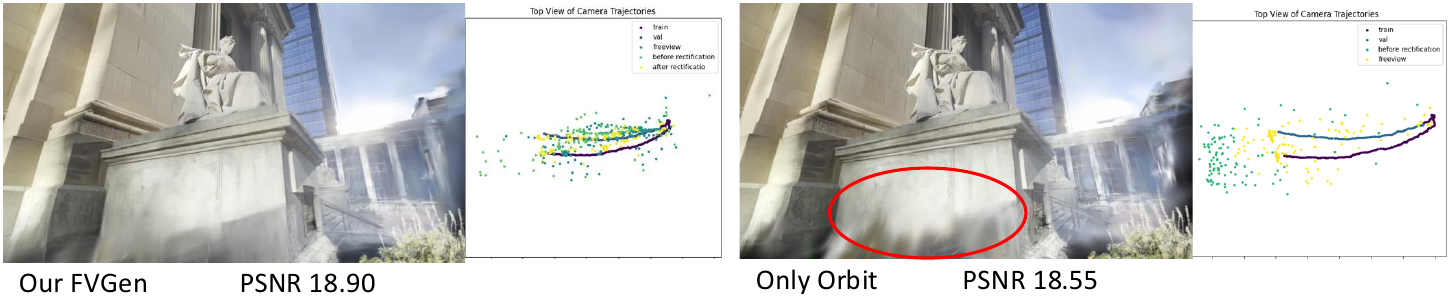}
\label{fig:rebuttal-inconsistency}
\vspace{-1em}
\caption{\textbf{Impact of diverse camera trajectory modes.} Relying solely on an \texttt{Orbit} trajectory limits viewpoint diversity, leading to noticeable blurring artifacts in under-observed regions (red circle). In contrast, our multi-mode sampling ensures maximal scene coverage, yielding sharper structural details and improved quantitative performance.}
\label{fig:appendix-camera}
\end{figure*}

\begin{figure*}[htbp]
\centering
\includegraphics[width=1\textwidth]{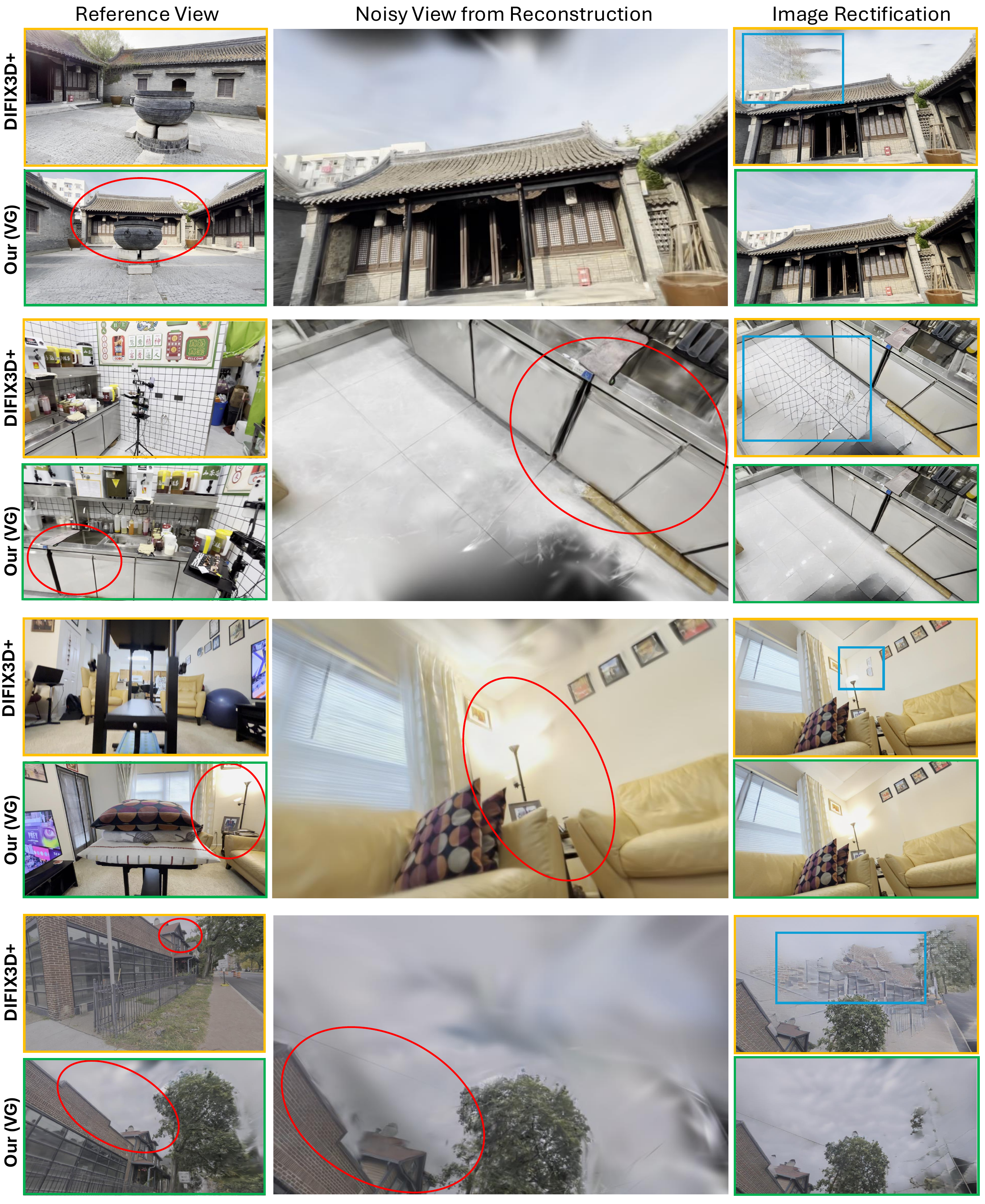}
   \vspace{-4mm}
   \caption{\textbf{Additional showcases of view graph impact on reference image selection.} The red circles highlight the shared visible region between the reference view and sampled noisy view, while the blue bounding boxes delineate artifacts introduced by inaccurate reference images during the image rectification stage.
   }
\label{fig:supp-refimage}
\end{figure*}

To guarantee comprehensive scene coverage, our framework initializes candidate viewpoints using a diverse set of trajectory modes, relying on the certainty-based view graph to efficiently filter out spatial redundancy. As shown in Figure~\ref{fig:appendix-camera}, compared to relying solely on the \texttt{Orbit} trajectory, incorporating diverse modes enables the data engine to capture geometrically challenging and under-constrained regions. This strategy not only significantly boosts overall synthesis performance but also ensures that the framework remains highly robust to the specific design choices of the candidate pool.

\begin{figure*}[htbp]
\centering
\includegraphics[width=1\textwidth]{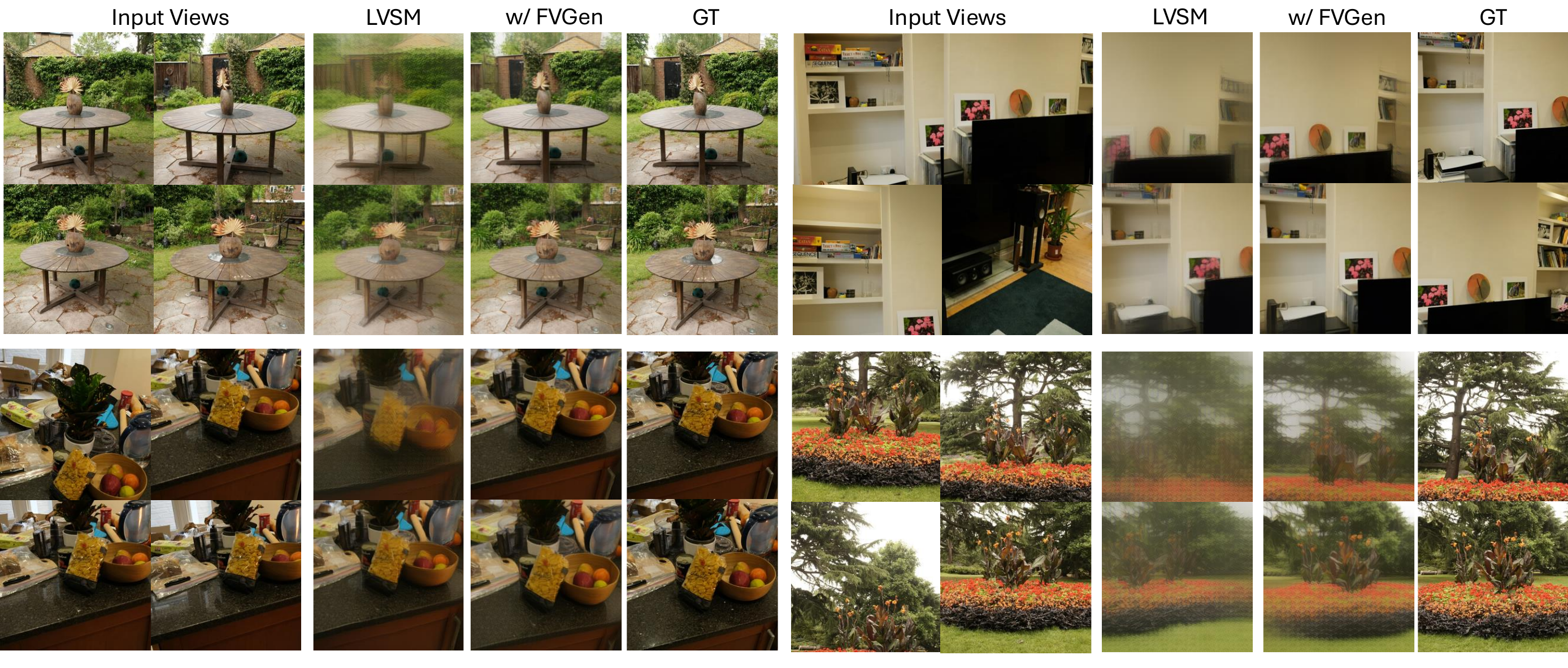}
   \vspace{-4mm}
   \caption{\textbf{Qualitative comparison of feed-forward on out-of-domain data (MipNeRF360).} The results are from LVSM at resolution 256.
   }
\label{fig:supp-ffood}
\end{figure*}

\section{Additional Qualitative Results}\label{sec:supp_results}
In this part, we provide more qualitative comparison for feedforward model and per-scene reconstruction.

\begin{figure*}[htbp]
\centering
\includegraphics[width=1\textwidth]{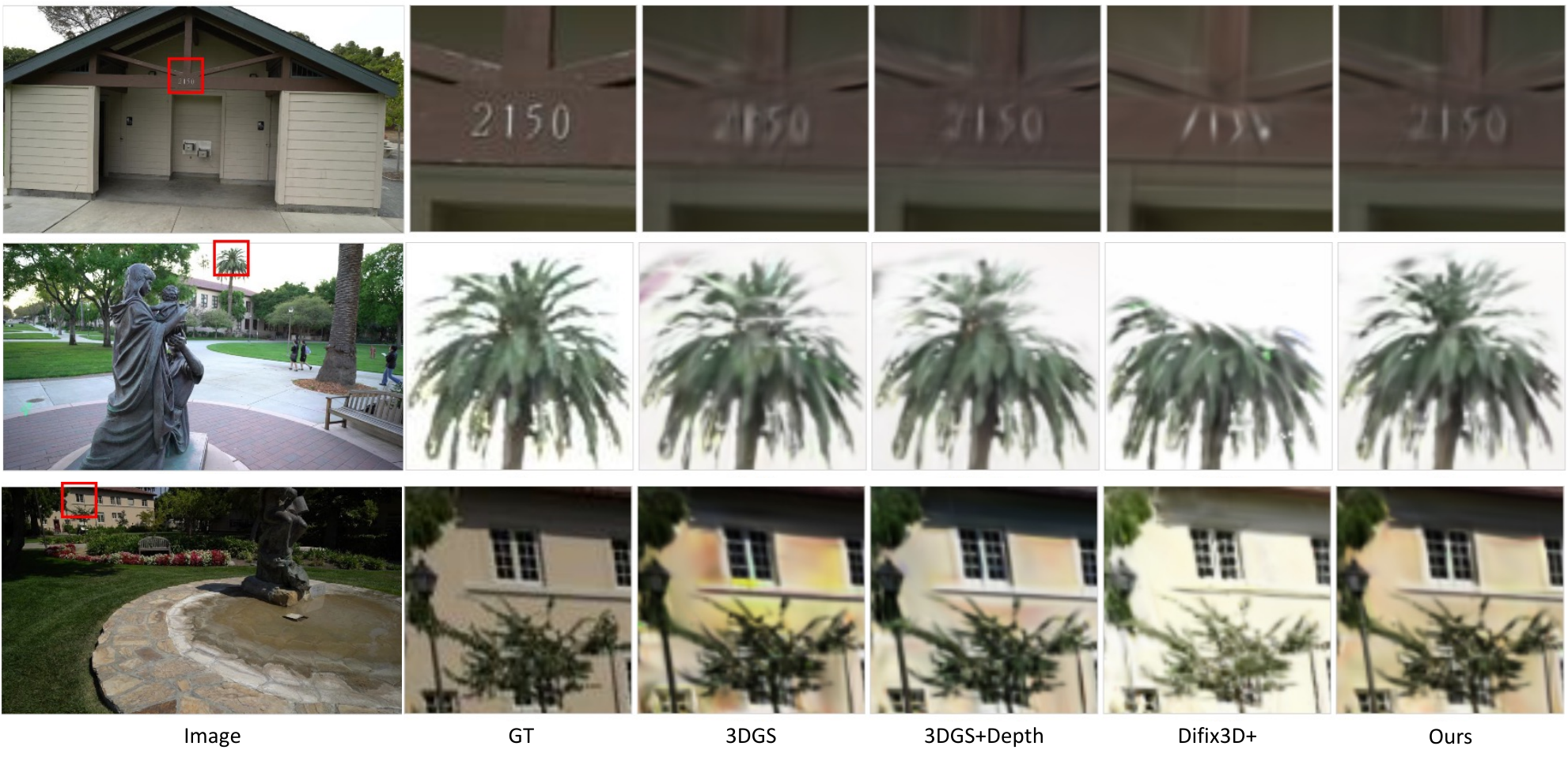}
   \vspace{-4mm}
   \caption{\textbf{Qualitative comparison of per scene reconstruction on Tanks and Temples dataset.}
   }
\label{fig:supp-tnt_perscene}
\end{figure*}

\begin{figure*}[h]
\centering
\includegraphics[width=1\textwidth]{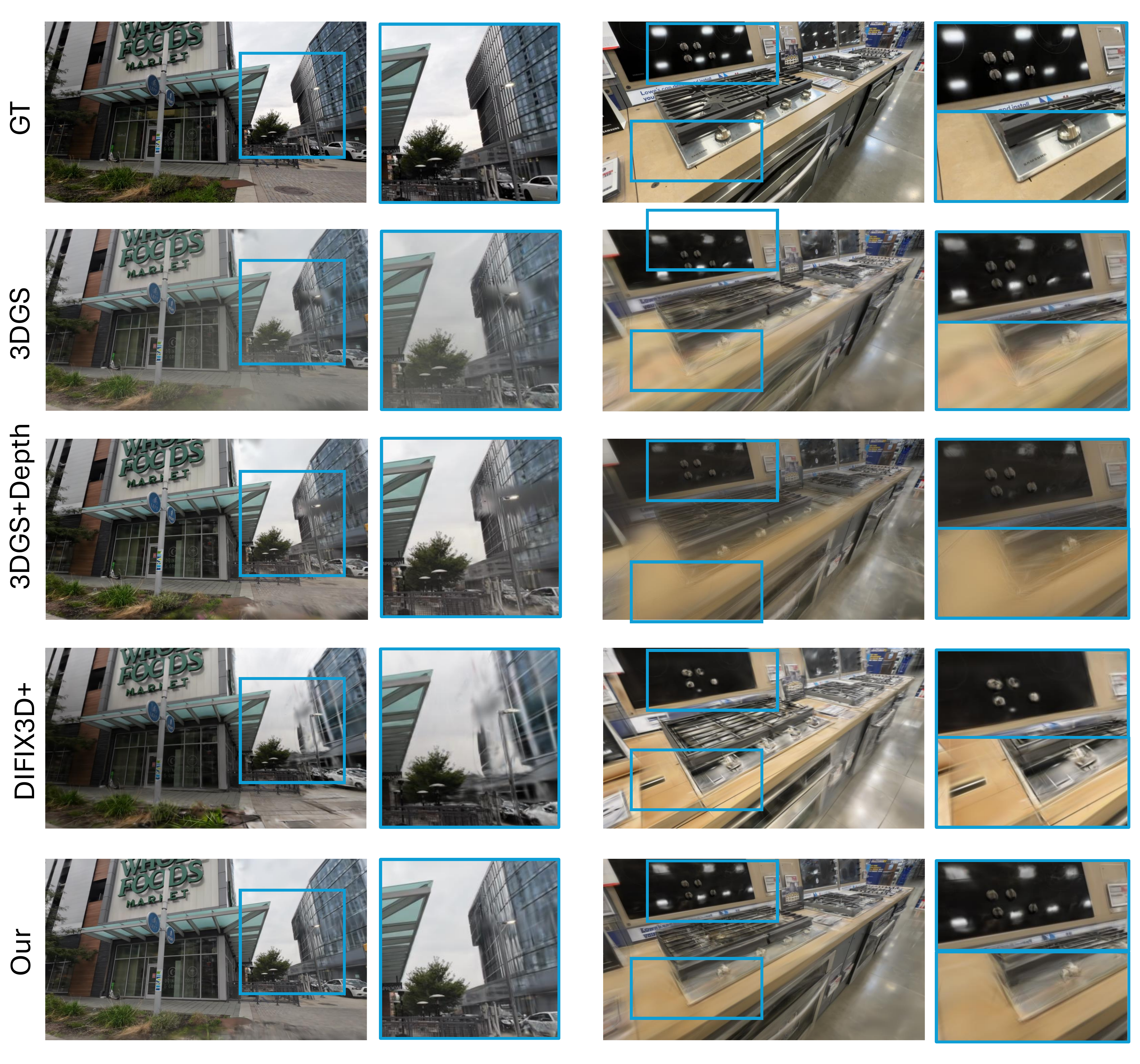}
   \vspace{-4mm}
   \caption{\textbf{Qualitative comparison on DL3DV dataset for per-scene reconstruction.} The blue bounding box indicates the zoom-in area. Despite the apparent clarity of DIFIX3D+, its progressive update and inaccurate reference image selection introduce significant hallucinated content, visible in the spurious reflection of the lamp and the corrupted desktop details on the right side.
   }
\label{fig:supp-dl3dv_perscene1}
\end{figure*}

\begin{figure*}[htbp]
\centering
\includegraphics[width=1\textwidth]{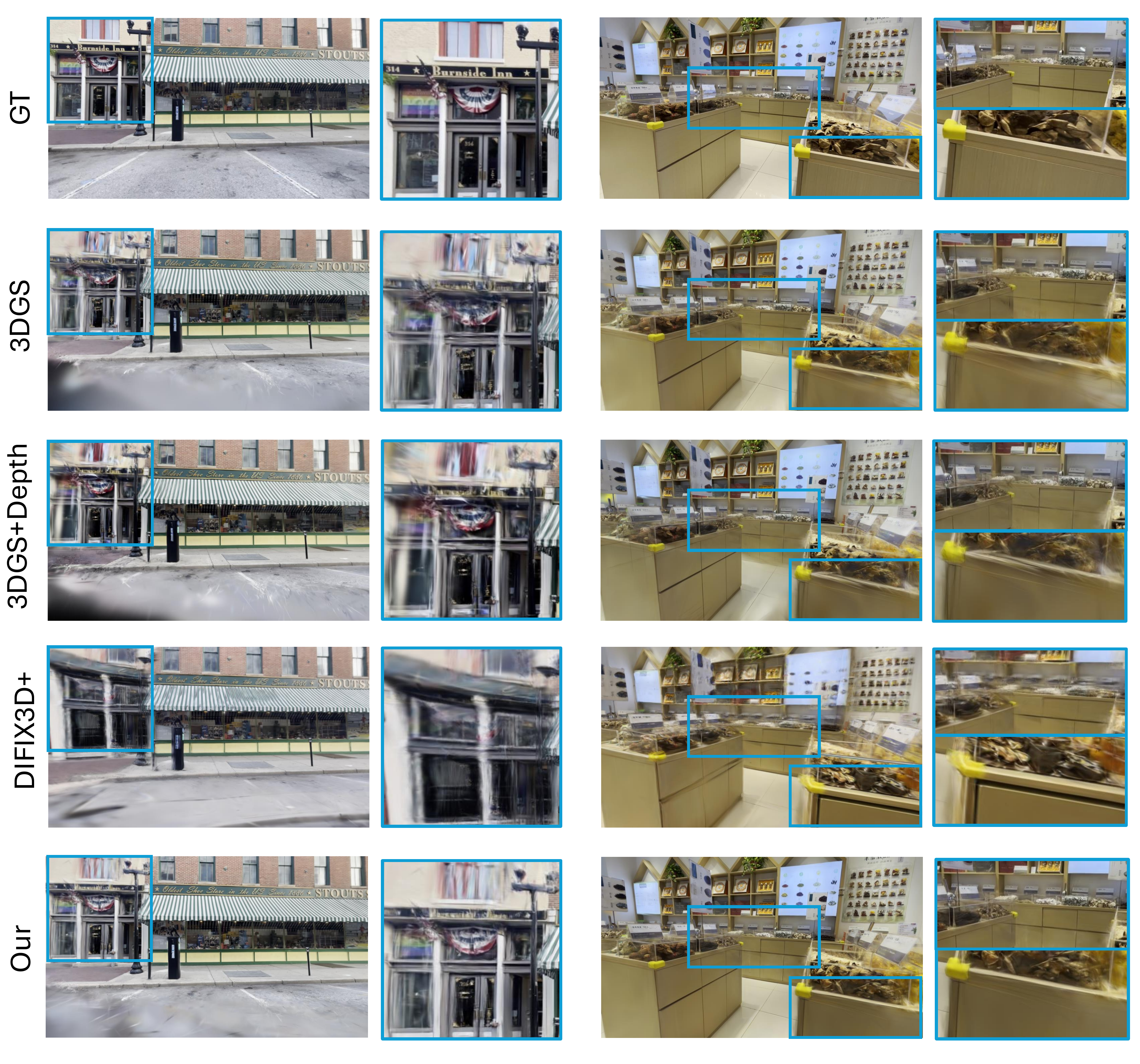}
   \vspace{-4mm}
   \caption{\textbf{Qualitative comparison on DL3DV dataset for per-scene reconstruction.} The blue bounding box indicates the zoom-in area.
   }
\label{fig:supp-dl3dv_perscene2}
\end{figure*}

\begin{figure*}[htbp]
\centering
\includegraphics[width=1\textwidth]{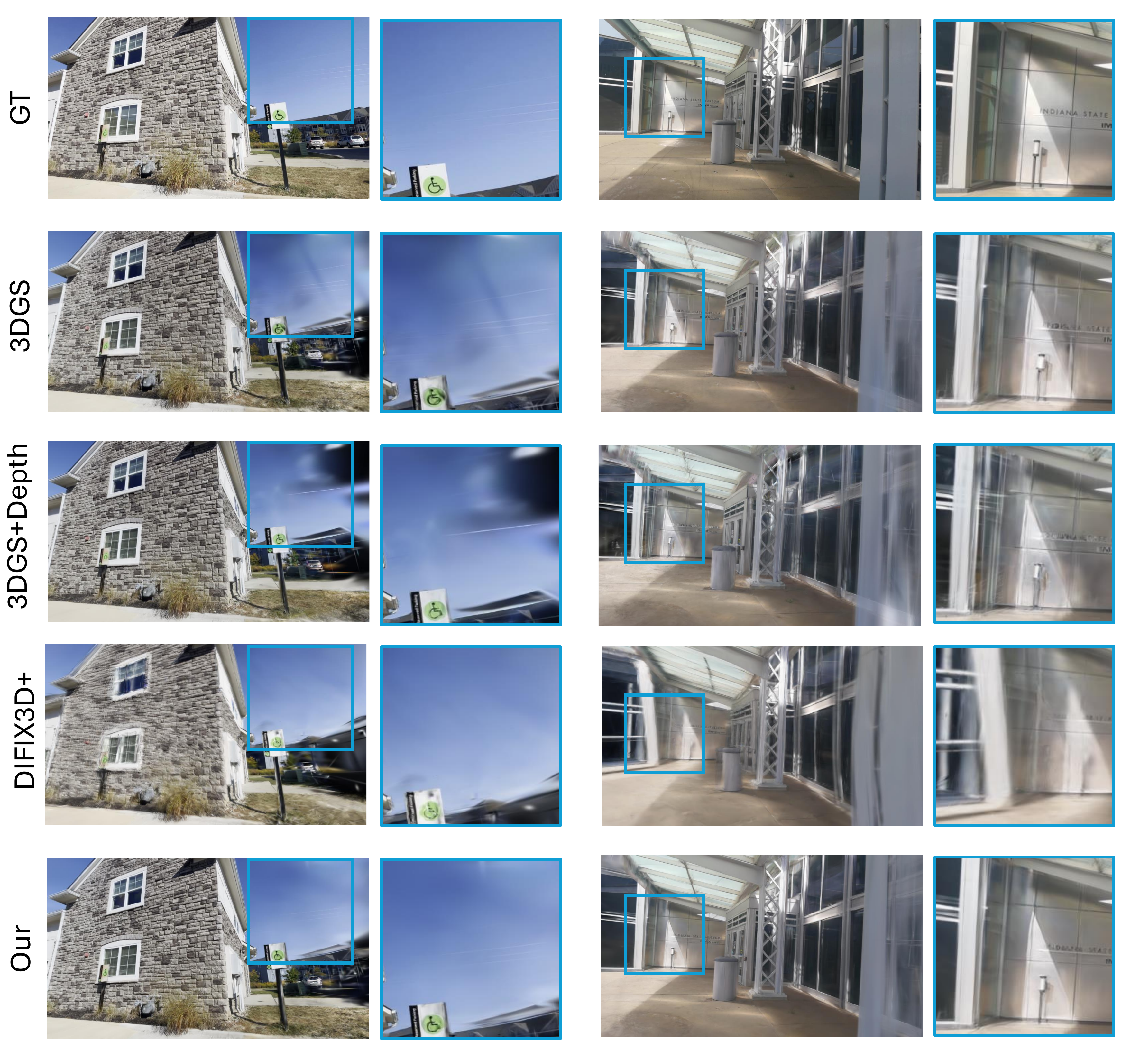}
   \vspace{-4mm}
   \caption{\textbf{Qualitative comparison on DL3DV dataset for per-scene reconstruction.} The blue bounding box indicates the zoom-in area.
   }
\label{fig:supp-dl3dv_perscene3}
\end{figure*}


\vspace{2pt}
\noindent\textbf{Out-of-Domain Results of FeedForward Model.}
We provide a qualitative comparison of the feed-forward model performance on out-of-domain (OOD) data, specifically using the MipNeRF360 dataset. Figure~\ref{fig:supp-ffood} illustrates the novel views generated by the baseline LVSM model against our approach incorporating FVGen for scaling scene data, benchmarked against the ground truth (GT). The results are rendered at a resolution of 256. The comparison highlights the advantages of FVGen: Unlike the baseline output, which suffers from excessive blurriness for OOD scenes, augmenting the training set with FVGen significantly mitigates this issue, producing sharper results closer to the GT.

\noindent\textbf{Per-scene Reconstruction. }
We show a per-scene reconstruction comparison for the Tanks \& Temples dataset in Figure~\ref{fig:supp-tnt_perscene}. And results on DL3DV dataset can be found in Figure~\ref{fig:supp-dl3dv_perscene1},~\ref{fig:supp-dl3dv_perscene2} and~\ref{fig:supp-dl3dv_perscene3}.

\section{Limitation and Future Works}\label{sec:supp_limitation}
The primary limitation lies in the Free-View Rectification stage, as the final image quality depends on the external diffusion model used for enhancement. Despite our certainty-aware View Graph improving reference image selection by ensuring geometric correspondence, residual artifacts can still be introduced. For future work, we plan to address this issue by fine-tuning the external diffusion model directly based on our sampling strategy, thereby reducing the synthetic-to-real domain gap. Integrating the view-specific certainty visibility mask into the diffusion model's conditioning. This would explicitly guide the denoiser to prioritize refinement only in uncertainty regions, thereby preventing artifacts in the original image distribution.

\begin{figure*}[h]
\centering
\includegraphics[width=1\textwidth]{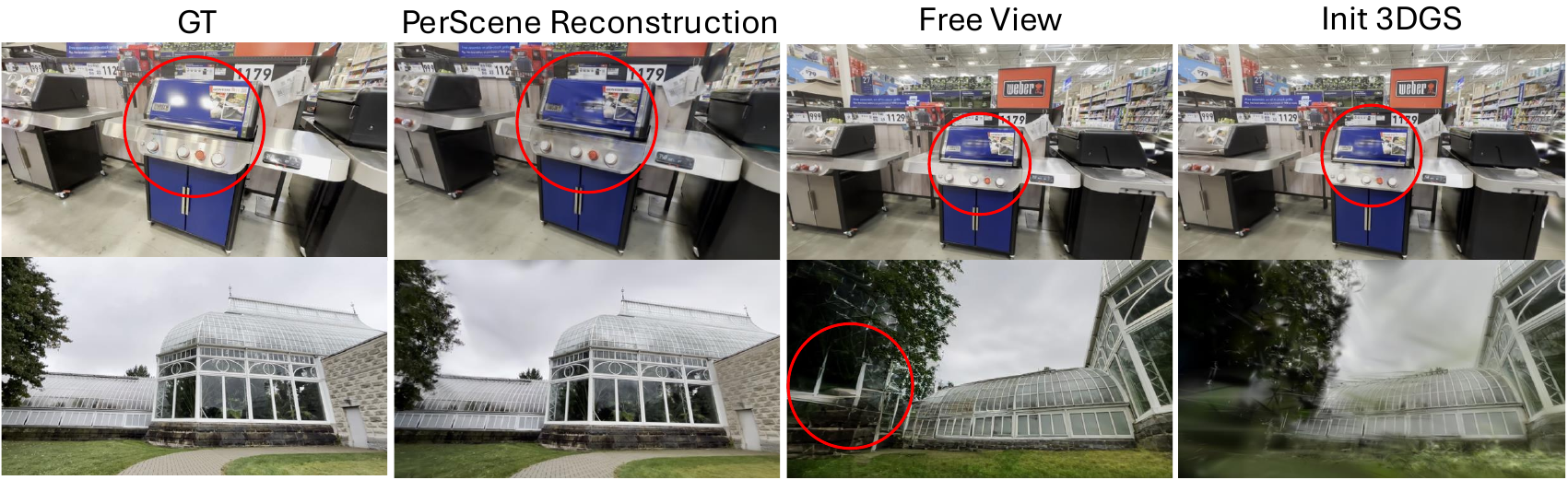}
\caption{\textbf{Failure cases of free-view generation.} Red circles indicate regions where our pipeline struggles due to diffusion priors, including the incorrect handling of complex reflections (top row) and the over-sharpening of 3DGS floaters (bottom row).}
\label{fig:appendix-failure}
\vspace{-1em}
\end{figure*}

\paragraph{Failure Cases and Limitations.}
The failure cases and applicability boundaries of our generated free-views primarily stem from two factors:
(1) \textbf{Diffusion Model Limitations:} Because the diffusion refinement model is trained for deblurring, it struggles to correct complex, view-dependent reflections as illustrated in Figure~\ref {fig:appendix-failure} (red circle). Furthermore, it can occasionally misinterpret severe 3DGS floaters from the initial reconstruction as valid scene structures, resulting in over-sharpened artifacts.
(2) \textbf{Free-View Scarcity:} In extreme conditions, such as extreme low-light environments, our rigorous quality filtering mechanism may reject a large number of poor renderings, leading to a scarcity of valid free-views.
Despite these localized limitations, downstream feed-forward NVS models trained with our augmented data exhibit strong robustness; they effectively treat these inconsistent artifacts as noise, thereby maintaining high overall synthesis performance.

\end{document}